\pdfoutput=1

\documentclass[11pt]{article}

\usepackage[final]{latex/acl}
\usepackage{times}
\usepackage{setspace} 
\usepackage{latexsym}
\usepackage{multirow}
\usepackage{subcaption}
\usepackage{CJKutf8}
\usepackage{}
\usepackage{float}
\usepackage{csquotes}
\usepackage[T1]{fontenc}
\usepackage{booktabs,multirow}
\usepackage{siunitx}

\usepackage[utf8]{inputenc}

\usepackage{microtype}

\usepackage{inconsolata}
\usepackage{tabularray}
\UseTblrLibrary{booktabs}
\usepackage{arydshln}
\usepackage{amssymb}
\usepackage{pifont}
\usepackage{xcolor}
\usepackage{amsmath}
\usepackage{pgfplots}
\pgfplotsset{compat=1.18}
\usepackage{array}
\usepackage{caption}
\usepackage{here}

\usepackage{graphicx}
\title{Understanding and Controlling Repetition Neurons and Induction Heads in In-Context Learning}
\author{
  Nhi Hoai Doan \quad\quad Tatsuya Hiraoka \quad\quad Kentaro Inui \\
  Mohamed bin Zayed University of Artificial Intelligence (MBZUAI)\\
  RIKEN\\
  \texttt{\{nhi.doan, tatsuya.hiraoka, kentaro.inui\}@mbzuai.ac.ae}
}

\begin{document}
\maketitle
\begin{abstract}
This paper investigates the relationship between large language models’ (LLMs) ability to recognize repetitive input patterns and their performance on in-context learning (ICL).
In contrast to prior work that has primarily focused on attention heads, we examine this relationship from the perspective of skill neurons, specifically repetition neurons.
Our experiments reveal that the impact of these neurons on ICL performance varies depending on the depth of the layer in which they reside.
By comparing the effects of repetition neurons and induction heads, we further identify strategies for reducing repetitive outputs while maintaining strong ICL capabilities~\footnote{Code: \url{https://github.com/hnhine/repnr_ind}}.
\end{abstract}

\section{Introduction}

Recent advances in LLMs have led to substantial performance improvements across a wide range of tasks through in-context learning (ICL). ICL can be understood as the model’s capability to identify repeated patterns within few-shot demonstrations and to learn to perform the target task based on those patterns~\cite{brown_icl}. Despite these advances, repetitive generation (i.e., where models produce overly redundant outputs) remains a fundamental challenge that can degrade output quality and potentially impair ICL performance~\cite{chiang-chen-2021-relating,li2023repetition}. 

Prior research has identified induction heads as key attention mechanisms within LLMs, essential for recognizing repeated patterns that underlie their ICL capabilities~\cite{olsson2022incontextlearninginductionheads,bietti2023birth,dong-etal-2024-survey}. The experimental results in the prior work suggest that these attention heads might affect the repetitive outputs. However, removing these induction heads has been shown to severely impair ICL performance due to their critical role in pattern recognition~\cite{crosbie-shutova-2025-induction}. Another line of work has begun exploring neuron-level interventions, discovering specific repetition neurons whose deactivation reduces redundant generation~\cite{wang2024mitigating,hiraoka-inui-2025-repetition}. Nevertheless, prior research has primarily addressed repetition and ICL separately, lacking systematic analysis on how these neuron-level interventions affect ICL performance, particularly their interaction with induction heads.
\begin{figure}
    \centering
    \includegraphics[width=1.1\linewidth]{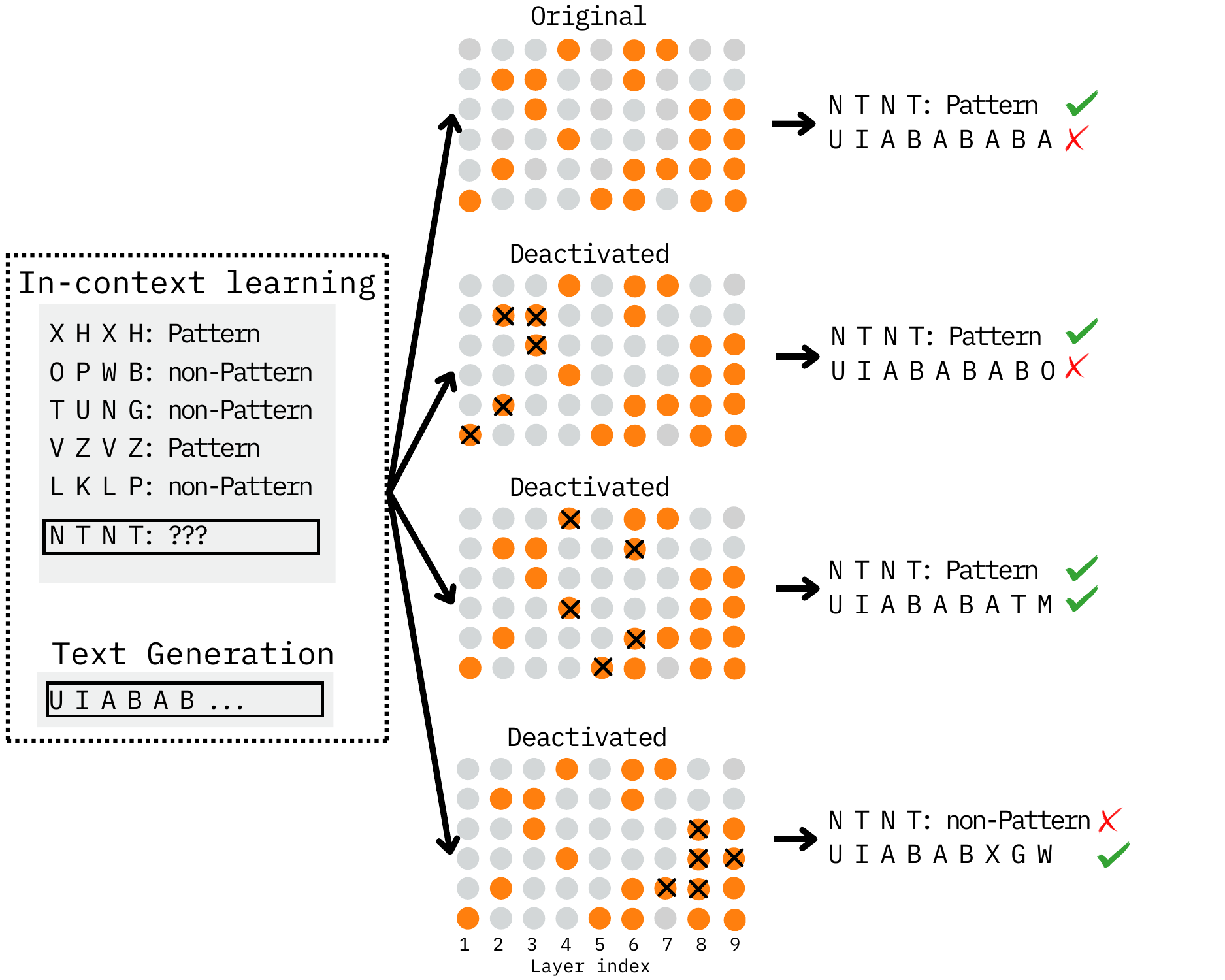}
    \caption{\textbf{Repetition neurons} (orange nodes), known for their strong activation in repetitive text generation, \textbf{play a causal role in few‐shot ICL}. Layer-wise ablation shows that deactivating: initial layers→ negligible effect on ICL or generation; middle layers→small drop in ICL recall and reduced repetition; last layers→severe ICL failure but degrades repetitive generation.}

    \label{fig:intro}
\end{figure}

Recent work by \citet{yan2024understanding} explored repetitive patterns' dual role (e.g., helping and hindering) in ICL but remained limited to sequence-level analyses. Similarly, \citet{analyzing-mitigating-repetitions} proposed strategies to mitigate repetitive loops during text generation without examining the impact on underlying ICL mechanisms. Thus, the interaction between repetition neurons and induction heads, and the potential for neuron-level control over ICL performance, remains largely unexplored.

To address these gaps, we present the first comprehensive layer-wise analysis examining how repetition neurons influence both repetitive generation and ICL performance(\S\ref{sec:layer_wise}, Figure \ref{fig:intro}). Our main contributions are: (1) We demonstrate that the impact of neuron ablation varies significantly with layer depth, supporting our hypothesis that neurons offer more granular control than attention heads, (2) Repetition neurons in output-side layers serve as critical executors of pattern-based predictions, exhibiting effects on ICL performance comparable to induction heads; (\S\ref{sec:com_ab}), and (3) Joint ablation of both components leads to a near-complete breakdown of pattern recognition (\S\ref{sec:joint_ab}). Crucially, we propose a novel three-segment ablation strategy that selectively deactivates repetition neurons in middle layers, reducing repetition while preserving ICL performance (\S\ref{sec:text_gen}). This approach offers a practical solution to mitigate the trade-off observed in prior work, providing finer control over LLM behavior compared to head-level interventions.

\section{Related Work}
Prior mechanistic analyses of LLM behavior have pursued two complementary strands: attention-head studies that illuminate pattern matching, and neuron-level investigations that target repetitive generation.
\subsection{Induction heads for pattern matching}
To pinpoint the specific attention heads that drive in-context pattern matching, we adopt a purely task-agnostic, prefix-matching probe. Initially, \citet{elhage2021mathematical} introduced the concept of induction heads, which refers to attention heads that have the ability to capture previous context by prefix-matching and copying mechanisms. Later on, \citet{olsson2022incontextlearninginductionheads} showed that these heads emerge abruptly during an induction‑bump phase early in training and that ablating them wipes out most in‑context learning. Following this work, \citet{bansal-etal-2023-rethinking} found that when pruning heads in ascending order of their copying scores, the total copying capacity falls off almost linearly from the first heads removed. It suggests that copying ability is distributed across nearly all heads. Therefore, we argue that a high prefix-matching score alone robustly flags “induction heads,” while the complementary copying measure does not distinguish them as cleanly. This feature is identical to the \citet{crosbie-shutova-2025-induction} experiment.

\subsection{Decoding interventions and neuron-level controls}
Repetition in generation has predominantly been addressed via decoding strategies such as nucleus sampling \cite{holtzman2020curiouscaseneuraltext} and top-k sampling \cite{fan-etal-2018-hierarchical}. These methods focus solely on model output and neglect the internal mechanisms responsible for repetitive outputs. Recently, neuron-level studies have identified specific repetition neurons responsible for redundancy in generation \cite{wang2024mitigating,hiraoka-inui-2025-repetition}. Yet, these studies did not examine interactions with other internal mechanisms like induction heads, crucial for pattern-based ICL.

\citet{yan2024understanding} and \citet{analyzing-mitigating-repetitions} separately explored repetition’s relationship with ICL and generation loops, respectively, but neither addressed neuron-head interactions nor conducted systematic, layer-wise neuron analyses. Our work bridges these research threads, conducting the first joint mechanistic exploration into how repetition neurons and induction heads collectively shape the trade-off between repetitive generation and ICL. We extend beyond existing literature by systematically investigating layer-wise neuron functionality, thereby providing practical strategies for targeted interventions that balance repetition control and robust ICL performance.


\section{Experimental Setting}
In this paper, we investigate the impact of the repetition neurons and induction heads on the ICL performance by ablating them.

\subsection{Models}
We evaluate mainly on Llama-3.1-8B \citep{grattafiori2024llama3herdmodels} because prior studies have reported similar inner workings regarding the repetition problem across various LLMs.
The specification of the Llama-3.1-8B model, with 32 layers, 458,752 neurons, and 1,024 heads in total, can be considered as a typical architecture for a model with this parameter scale.
For the comparison among the models,  Appendix \ref{apx:lw_repnr}, \ref{apx:ab_id} and \ref{apx:joint_ab} shows the extended results on Qwen2.5-7B \citep{qwen2025qwen25technicalreport} and LlaMA-2-13B \citep{touvron2023llama}. 

\subsection{Datasets}
\label{sec:datasets}
\begin{table}[t]
\small
\centering
\begin{tabular}{@{}lcc@{}}
\toprule
\textbf{Task} & \textbf{Pattern} & \textbf{Non-Pattern} \\
\midrule
Repetition &
\(\begin{aligned}
\text{X H X H}
\end{aligned}\)
&
\(\begin{aligned}
\text{Z E W F}
\end{aligned}\) \\

Recursion &
\(\begin{aligned}
\text{V D D D}
\end{aligned}\)
&
\(\begin{aligned}
\text{N O W T}
\end{aligned}\) \\

Centre-embedding &
\text{X B V B X}

&
\(\begin{aligned}
\text{L Q I F P}
\end{aligned}\) \\

WordSeq 1 &
\(\begin{aligned}
\text{grape shark}\\
\end{aligned}\)
&
\(\begin{aligned}
\text{onion cello}
\end{aligned}\) \\

WordSeq 2 &
\(\begin{aligned}
\text{pumpkin scooter}
\end{aligned}\)
&
\(\begin{aligned}
\text{peach pilot}
\end{aligned}\) \\

\bottomrule
\end{tabular}
\caption{Each letter-sequence dataset features examples following the respective patterns labeled “Pattern” and random sequences labeled “Non-Pattern.” For word-sequence tasks, examples include pairs of semantically categorised words \cite{crosbie-shutova-2025-induction}.}
\label{tab:dataset}
\end{table}

To investigate the impact of ablating repetition neurons and induction heads on the ICL performance, we use five well-controlled synthetic datasets, following the prior work ~\citep{gao2021framework, crosbie-shutova-2025-induction}.
As shown in Table \ref{tab:dataset}, all tasks are binary classification for ``Pattern'' and ``Non-Pattern'' inputs.
For example, the task named ``Repetition'' is a binary classification of whether the four-letter input text has repetitive patterns (e.g., an input ``XHXH'' has a repetitive pattern)\footnote{\citet{crosbie-shutova-2025-induction} explains all the tasks in detail. We re-create the dataset by ourselves to use a larger number of samples.}.

For each of the Repetition, Recursion, and Centre-embedding tasks, we construct a dataset of 1,000 sequences, evenly divided into 500 Pattern and 500 Non-Pattern examples.
Each WordSeq task comprises 500 word‐pair examples, evenly split into Pattern and Non‐Pattern classes. 



For WordSeq 1, the Pattern set consists of 250 \(\langle\)fruit, animal\(\rangle\) pairs, while the Non-Pattern set contains 250 examples randomly drawn from 1000 pairs of the four relations  
\(\langle\)vegetable, vehicle\(\rangle\), \(\langle\)vegetable, instrument\(\rangle\), \(\langle\)body-part, vehicle\(\rangle\) and \(\langle\)body-part, instrument\(\rangle\). This design ensures that Non-Pattern inputs are semantically and structurally distinct, introducing noise that contrasts sharply with the homogeneous Pattern class. In WordSeq 2, we sample 500 pairs, but swap the target relation: the Pattern input is \(\langle\)vegetable, vehicle\(\rangle\), the Non-Pattern is from \(\langle\)fruit, animal\(\rangle\), \(\langle\)fruit, profession\(\rangle\), \(\langle\)furniture, profession\(\rangle\) and \(\langle\)furniture, animal\(\rangle\).

\begin{figure}
    \centering
    \includegraphics[width=1\linewidth]{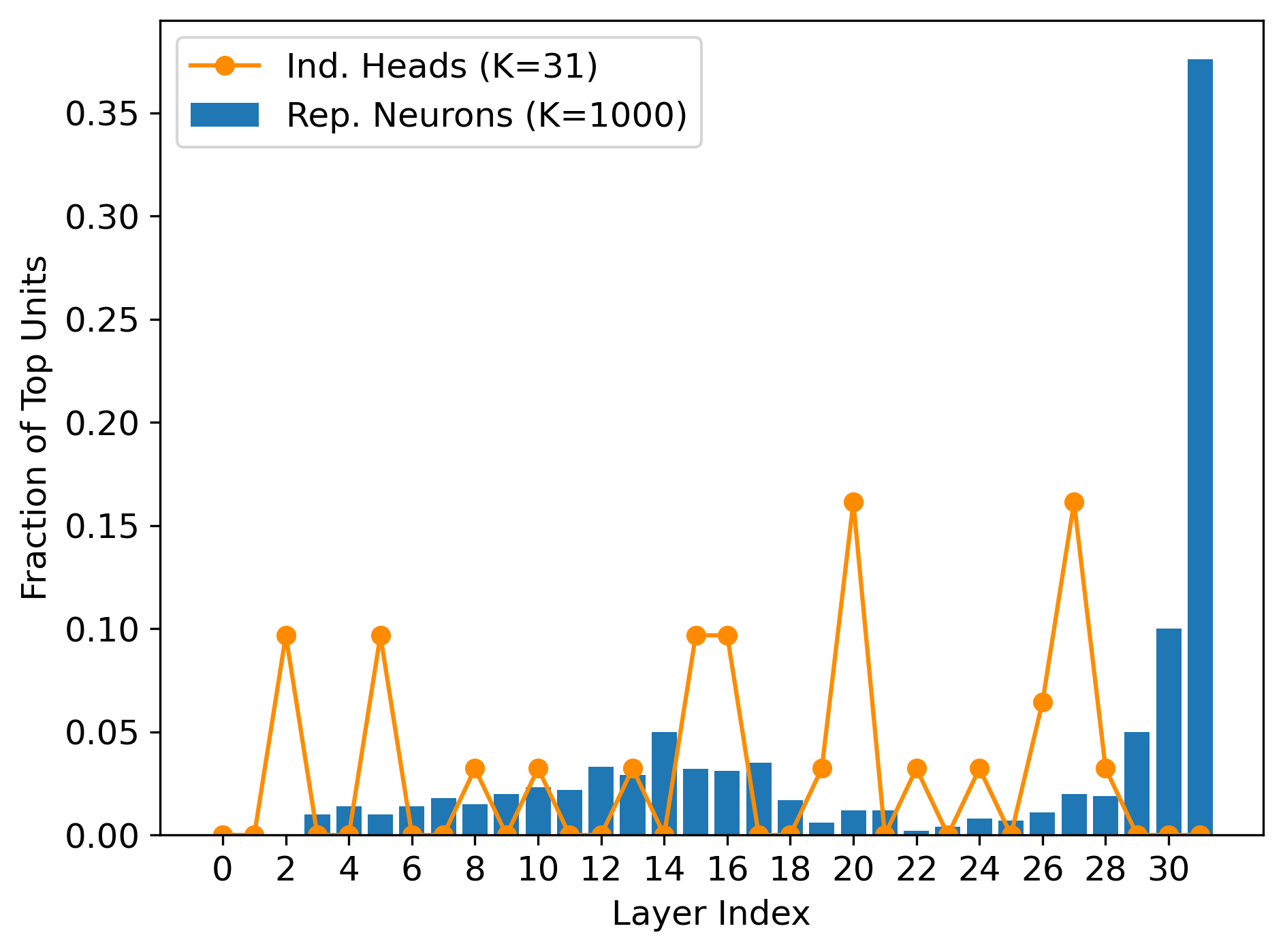}
    \caption{Distribution of top 31 (3\%) induction heads and top 1000 (0.2\%) repetition neurons across layer.}
    \label{fig:layer-comparison}
\end{figure}

\subsection{Repetition Neurons Identification}
\label{sec:neuron-identification}
We identify the repetition neurons to be ablated in our experiments with the activation-based method~\cite{hiraoka-inui-2025-repetition}.
This method finds repetition neurons by focusing on the difference between the model's activation when generating normal text and repetitive text.

Once the repetition onset $s$ is marked on a text, we extract each feed-forward (FFN) neuron’s activation over a window of $r=30$ tokens before ($[s-r, s-1]$) and after ($[s, s+r-1]$) the onset. Averaging these activations across all texts in a group yields ``normal'' and ``repetition'' means, $a_n$ and $\bar a_n$, respectively, for each neuron $n$. We then compute
\[
\Delta_n = \bar a_n - a_n
\]
and rank neurons by descending $\Delta_n$ to obtain the top-$K$ neurons as repetition neurons. A higher difference value $\Delta_n$ suggests the neuron triggers the model to produce an error. For this identification process, we prepared 1000 texts~\footnote{See details in Appendix \ref{apx:repnr}} that include repetitive phrases from each LLM's generation.

\subsection{Induction Heads Identification}
\label{sec:head-identification}

The induction heads are identified based on the prefix-matching score~\cite{elhage2021mathematical}, which is calculated from the attention weight to tokens that are exactly the same as the one to be generated in the next time step.

Concretely, we randomly sample sequences of 50 tokens\footnote{To exclude the effect of high-frequency tokens, we ignored the 4\%  most and least frequent tokens in the vocabulary when sampling tokens, following the prior work~\cite{crosbie-shutova-2025-induction}.} and repeat each sequence four times.
Then we fed them to the model to calculate the prefix-matching score.
For every head, we compute the average attention weight that each token places on its immediately preceding occurrence in earlier repeats. 
By repeating this over five independent draws and averaging, we obtain a stable prefix-matching score per head and select those with the highest scores as our induction heads.

Figure \ref{fig:layer-comparison} shows the distribution of the repetition neurons and induction heads per layer. 
Note that the repetition neurons and the induction heads are identified independently from the ICL dataset created in \S \ref{sec:datasets}.

\subsection{Evaluations}

Our primary purpose of the experiments is to investigate the effect of repetition neurons and induction heads on the performance of ICL tasks. 
To quantify their contributions, we compare the recall of the model on binary classification tasks before (original) and after ablating these neurons and heads.
In Sections~\S\ref{sec:com_ab} and~\S\ref{sec:joint_ab}, we define $\Delta\mathrm{Recall}$ as the difference between the model’s recall after intervention and its original recall.

For the fine-grained analysis of the impact of ablation on the ICL tasks, we separately report the recall on Pattern and Non-Pattern subsets, rather than the metric that combines them, such as the F1 score.
Prior works suggest that the repetition-related components play several roles, such as capturing the repetitive input and stimulating repetitive outputs~\cite{olsson2022incontextlearninginductionheads, yan2024understanding, hiraoka-inui-2025-repetition}.
Considering their ability to capture repetitive inputs, we would see a different trend in the performance change between the Pattern and Non-Pattern subsets when ablating these components.
Specifically, we hypothesize that they will disproportionately support correct labeling of Pattern examples, where the few-shot context itself repeats the target sequence, while having little impact on Non-Pattern inputs. We evaluated these subsets separately to identify the effect of neurons and heads on repetitive and non-repetitive samples. 

For all experiments, we used 10-shot examples for the ICL tasks if not otherwise mentioned.
We consider the token output immediately after the input query as the answer of the model\footnote{We used unrelated labels ``Foo'' and ``Bar'' instead of ``Pattern'' and ``Non-pattern,'' following \citet{crosbie-shutova-2025-induction}.}, and calculate the recall based on this answer. Each experiment is run three times independently, and we report the average result across these runs.

\section{Layer-wise Repetition Neuron Ablation}
\label{sec:layer_wise}
\begin{figure*}[t]
    \centering
    \includegraphics[width=1\linewidth]{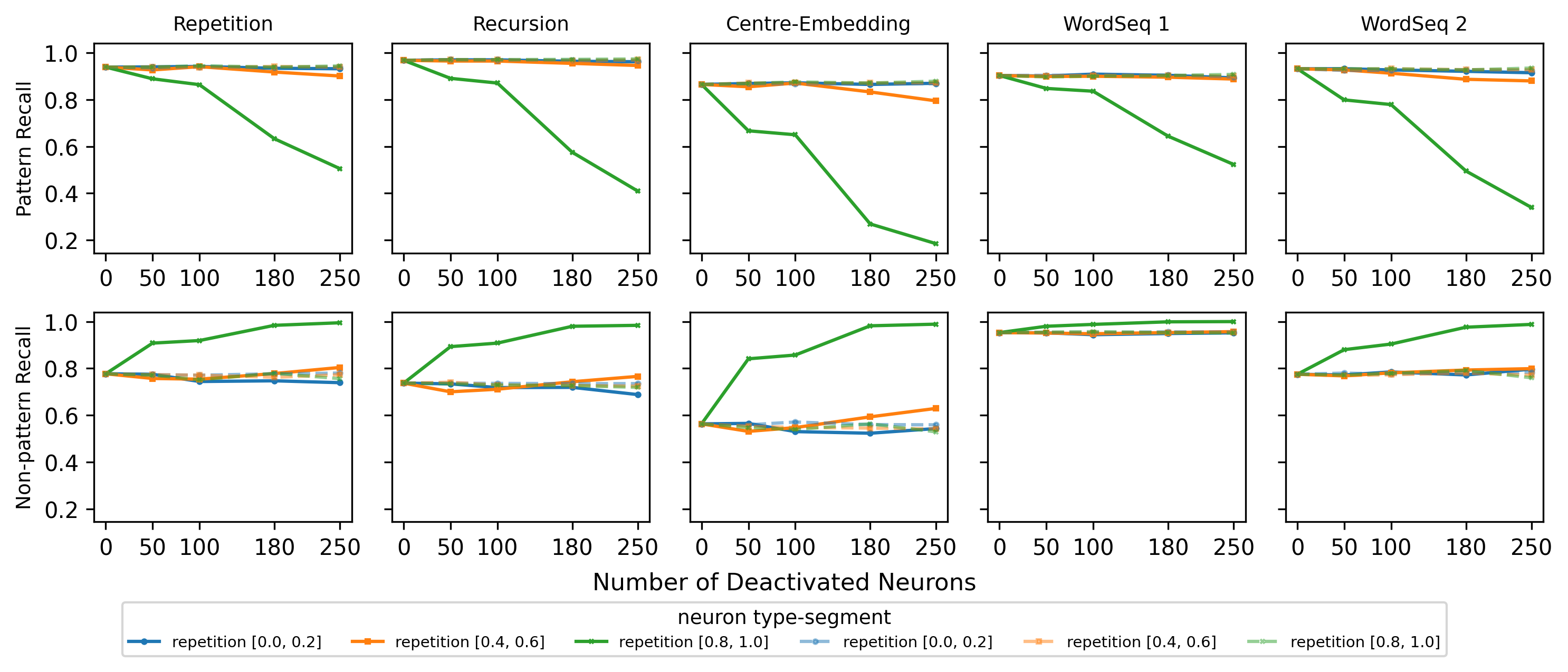}
    \caption{Model performance after ablating repetition neurons (solid lines) versus randomly selected neurons (dashed line) by layer-wise with the 10-shot ICL setting.
    The performance is separately reported for the Pattern (top) and Non-Pattern (bottom).
    Data points with 0 on the X-axis show the performance without ablation.
    }
    \label{fig:ab_repr_layer}
\end{figure*}
Building on our identification of repetition neurons (\S \ref{sec:neuron-identification}), we first ablate them layer-segment–wise to quantify their causal role in Pattern and Non-Pattern ICL performance.

\subsection{Neuron Segmentation}
To understand how repetition neurons drive few-shot ICL, we first locate these neurons within the model and quantify the contributions of different repetition neuron groups. 
The distribution across layers of repetition neurons in the model, as in Figure \ref{fig:layer-comparison}\footnote{More results with various numbers of neurons are shown in Figure \ref{fig:rep-dist}.}, is unbalanced. 

While it is consistent with previous studies on a higher number of task-specific repetition neurons in early layers, we are motivated by their position due to the high concentration of repetition neurons in late layers.
This pattern aligns with prior work showing that task-specific neurons usually appear in late transformer layers \citep{wang-etal-2022-finding-skill,hiraoka-inui-2025-repetition}. 
Because early-layer neurons project directly into the output logits, their concentration implies a stronger influence on model predictions. 
This observation motivates our targeted-ablation experiments, which quantify the causal contribution of repetition neurons in each layer segment to few-shot ICL.

To assess the causal contribution of repetition neurons to few-shot ICL, we partition the $L$-layer Transformer model into three depth-based segments and intervene on the neurons in each segment separately.
We compute the normalized depth of layer $i$ as $u_i = i/L$.
Layers whose $u_i$ falls in the ranges $0.0 < u_i < 0.2$, $0.4 < u_i < 0.6$, or $0.8 < u_i < 1.0$ are labeled as early, middle, and late layers, respectively.
For example, in a 32-layer model, layer 16 has $u_{16}=16/32=0.5$ and is assigned to the middle segment. 

By comparing ICL performance before and after interventions within each segment, we quantify the causal impact of repetition neurons across the model’s depth.
This segmentation ensures that our results generalize across the model and capture the distinct contributions of neurons from each part of the network.


\subsection{ICL Performance with Neuron Ablation}

Figure~\ref{fig:ab_repr_layer} reports the effect of ablating repetition neurons by layer segment in the 10-shot setting\footnote{We report the result of ablating up to 250 neurons because we can obtain the highest impact with 250. Figure \ref{fig:ab-nr-500-l38b} shows the performance change with ablating more neurons.}.
For ablating repetition neurons, we straightforwardly modify their activation value to $0$. Removing the top‐$K$ repetition neurons in the earlier and middle layers produces only a modest reduction in Pattern‐query recall, whereas deactivating those in the late segment layers drives a drop of up to 60\%.  This result runs counter to the “two-peak” hypothesis of \citet{hiraoka-inui-2025-repetition}, which posits that intermediate layers primarily detect repetition while final layers merely replicate it. Instead, our findings suggest that late-segment neurons play the dominant role in consolidating the repeated structures that underlie few-shot ICL. Mid-segment neurons do appear to contribute to repetition, but their function can be largely compensated for by downstream layers during ICL tasks. We hypothesize that, in the context of demonstrations already rich in repeated patterns, early detectors are less critical, and that the final layers are where pattern recognition is ultimately consolidated. In \S\ref{sec:text_gen}, we further explore how mid-layer neurons uniquely influence the generation of repetitive phrases, roles that cannot be fully rescued by later layers alone.

Moreover, ablating repetition neurons also affects performance on Non-Pattern inputs. By removing these neurons, the model becomes less biased toward detecting repetition and thus better distinguishes genuinely non‐repetitive examples, improving its ability to identify the correct class when no pattern is present. These observations suggest that repetition neurons contribute to capturing repetitive input even though these neurons are detected independently from the ICL tasks. Furthermore, we can conclude that neurons in different depths of layers have different impacts on the ICL performance.

\section{Attention Head Ablation}
\label{sec:com_ab}
\begin{figure*}[t]
    \centering
    \includegraphics[width=1\linewidth]{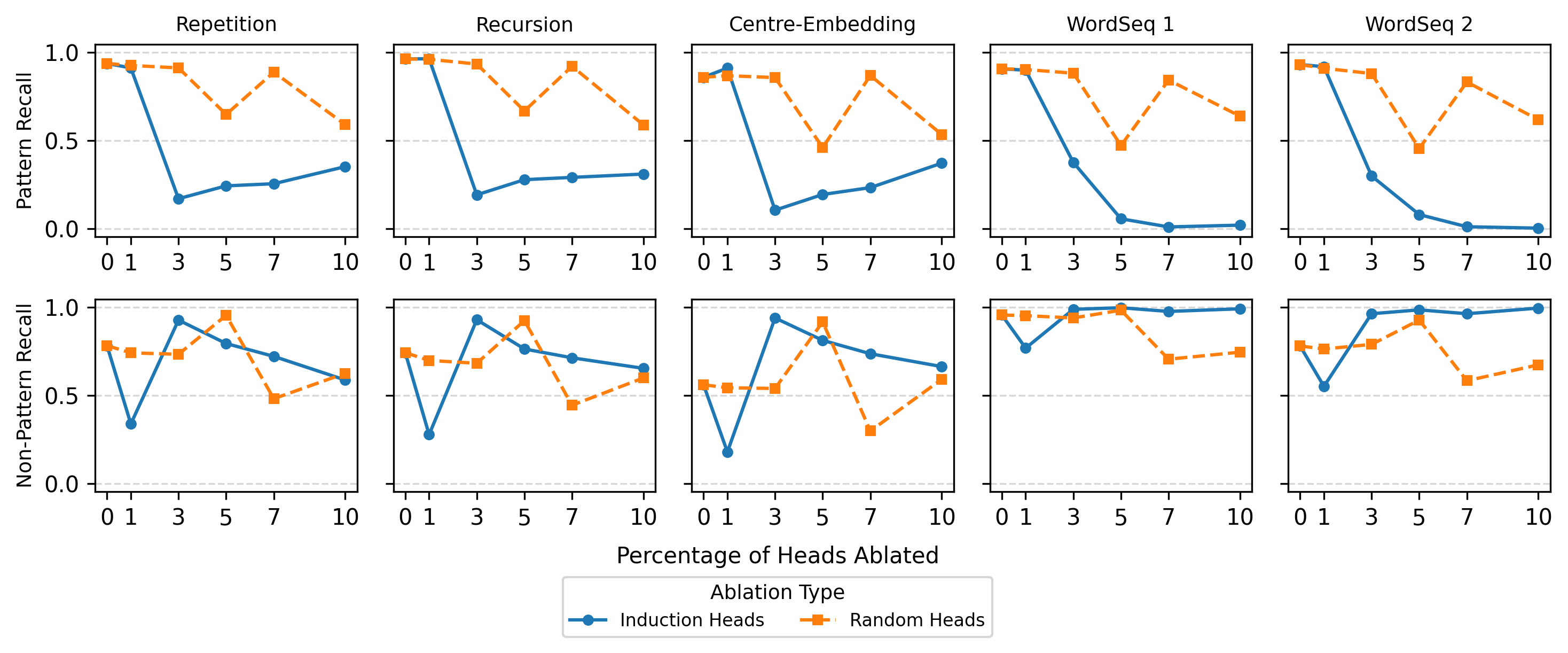}
    \caption{Llama-3.1-8B performance on the Pattern (top) and Non-pattern (bottom) subsets after ablating induction heads with the top highest prefix-matching scores on two classes in the 10-shot setting.
    Data points with 0 on the X-axis show the performance without ablation.
    }
    \label{fig:ab-id-class-llama8-10shot}
\end{figure*}

In this section, we investigate the impact of ablating induction heads on the same dataset. After identifying them (\S\ref{sec:head-identification}) via prefix-matching scores, we ablate their attention weights by setting them to zero during inference. As in \S\ref{sec:layer_wise}, we do not segment induction heads due to their limited total number, and the definition of an induction head is layer-agnostic (Figure~\ref{fig:layer-comparison}).

Unlike prior work such as \citet{crosbie-shutova-2025-induction}, which aggregates performance across Pattern and Non-Pattern inputs, our class-disaggregated evaluation isolates the causal effect of induction heads on pattern recognition. This design, motivated by their role as detectors of repetitive structure, is essential for tracing how their removal impairs ICL and enables downstream analysis of repetition neuron activation.

\subsection{ICL performance with Head Ablation}
The experimental results in Figure \ref{fig:ab-id-class-llama8-10shot} show a similar trend in the performance difference to that observed in the neuron ablation. For the Pattern inputs, ablating only one induction head has a minimal impact on model performance. 
However, removing three or more heads causes a catastrophic collapse: recall falls by 75.4\%–76.7\%. 
The two WordSeq tasks exhibit a similarly severe decline, with recall reductions of 53.1\%–63.2\%.
These results confirm that a small subset of induction heads accounts for the majority of pattern‐recognition capability: ablating more than three of these heads effectively disables the model’s ability to detect the target patterns.

For the Non-Pattern inputs, there is no clear trend. 
On the surface-level tasks (i.e., Repetition, Recursion, and Center-embedding), ablating three induction heads positively affects the performance similarly to the trend shown in the neuron ablation (Figure \ref{fig:ab_repr_layer}).
However, as the number of heads increases, this effect gradually weakens.
After observing the model's output with intervention, we found broken outputs (e.g. invalid unicode sequences). This suggests removing induction heads easily breaks the Llama-3.1-8B~\footnote{This trend depends on the model architecture. For example, Qwen2.5-7B yields smoother results on the same experiment (Figure \ref{fig:ab-id-class-qwen-10shot}), while Llama-2-13B shows a similar trend (Figure \ref{fig:ab-id-class-llama13-10shot}) to Llama-3.1-8B.}, supporting our hypothesis that controlling the model with ablating attention head is difficult.
In contrast, on two WordSeq tasks, we observe trends that align with the results in Figure \ref{fig:ab_repr_layer}.
This implies that the model can recover the lack of some induction heads when solving semantic-level repetition tasks.

\subsection{Cascade of Head to Neuron}
Combining results from Figures~\ref{fig:ab_repr_layer} and~\ref{fig:ab-id-class-llama8-10shot}, we observe that both induction heads and final-segment repetition neurons are critical for Pattern recognition. To investigate the hypothesized cascade, where induction heads activate repetition neurons, we directly measured the post-activation values of repetition neurons when ablating induction heads. 
\begin{table*}[htbp!]
\centering
\footnotesize
\setlength{\tabcolsep}{5.5pt}
\begin{tabular}{llrrrrrrrrrr}
\toprule
\multirow{2}{*}{\textbf{Segment}} & \multirow{2}{*}{} &
\multicolumn{2}{c}{Repetition} &
\multicolumn{2}{c}{Recursion} &
\multicolumn{2}{c}{Centre-Embedding} &
\multicolumn{2}{c}{WSQ1} &
\multicolumn{2}{c}{WSQ2} \\
\cmidrule(lr){3-4}\cmidrule(lr){5-6}\cmidrule(lr){7-8}\cmidrule(lr){9-10}\cmidrule(lr){11-12}
 &  & rep\_nr & ran\_nr & rep\_nr & ran\_nr &
      rep\_nr & ran\_nr & rep\_nr & ran\_nr &
      rep\_nr & ran\_nr \\
\midrule
\multirow{2}{*}{[0.0, 0.2]} & ind & \textbf{-0.120} & 0.029 & \textbf{-0.177} & 0.029 & 0.000 & 0.000 & 0.000 & 0.000 & 0.000 & 0.000 \\
                          & ran & -0.001 & 0.027 & -0.036 & 0.024 & 0.085 & 0.011 & 0.005 & 0.033 & -0.002 & 0.033 \\
\addlinespace[2pt]
\multirow{2}{*}{[0.4, 0.6]} & ind & \textbf{-0.105} & 0.022 & \textbf{-0.177} & 0.029 & 0.023 & 0.014 & 0.013 & 0.011 & 0.014 & 0.012 \\
                            & ran & -0.053 & 0.020 & -0.036 & 0.024 & 0.059 & 0.028 & 0.020 & 0.025 & 0.030 & 0.028 \\
\addlinespace[2pt]
\multirow{2}{*}{[0.8, 1.0]} & ind & \textbf{-0.140} & 0.119 & \textbf{-0.138} & 0.117 & -0.023 & 0.058 & -0.026 & 0.056 & -0.029 & 0.056 \\
                            & ran & -0.035 & 0.034 & -0.031 & 0.023 & -0.015 & 0.044 & -0.013 & 0.052 & -0.009 & 0.044 \\
\bottomrule
\end{tabular}
\caption{\textbf{Activation change rate by segment and task (Llama-3.1-8B).}
Values are $\Delta=(\mathrm{mean}_{\text{ablated}}-\mathrm{mean}_{\text{normal}})/\mathrm{mean}_{\text{normal}}$ at the last prompt token (FFN gate post-activation).
\textbf{ind}: ablate top 3\% induction heads; \textbf{ran}: ablate 3\% random heads.
\emph{rep\_nr}: top-$K$ repetition neurons; \emph{ran\_nr}: $K$ random neurons from the same segment ($K{=}250$).}
\label{tab:ab_hd_avg_nr}
\end{table*}

Our findings also reveal a notable divergence in how different model families utilize induction heads. In Table \ref{tab:ab_hd_avg_nr} of Llama-3.1-8B, ablating induction heads primarily impairs structural pattern recognition (e.g., 10-17\% drop in Repetition, Recursion), while high abstract and semantic tasks (Centre-Embedding and WordSeq 1/2) exhibit greater robustness, suggesting the presence of alternative, non-induction pathways for semantic pattern encoding. In contrast, Qwen2.5-7B exhibits uniform sensitivity: the same ablation degrades performance and neuron activation across all tasks, including semantic ones (Appendix~\ref{apx:ab_id}, Tables \ref{tab:apx_abhd_avgnr_l8}-\ref{tab:apx_abhd_avgnr_l13}). This indicates that Qwen relies on induction heads not merely for surface-level repetition, but as a general-purpose mechanism for pattern abstraction at all levels.
We hypothesize that architectural or training differences, such as Qwen’s tokenizer or pretraining corpus, strengthen the coupling between prefix-matching and higher-level relational reasoning. In contrast to Llama, which appears to leverage alternative pathways for semantic tasks, Qwen relies more uniformly on induction heads for both structural and semantic pattern recognition. Thus, the functional scope of induction heads is not universal: it expands with model design, and interventions must be architecture-aware to preserve ICL.


Regarding the role of repetition neurons, we assume that they operate at different levels of granularity. Middle-layer repetition neurons appear to be tightly coupled with local and surface-level token replication. They are sensitive to exact token positions and local patterns. In contrast, induction heads, which operate via prefix matching, can encode relational patterns (e.g., ⟨fruit, animal⟩) that support abstract pattern recognition in ICL, even in semantic tasks. Therefore, ablating middle-layer repetition neurons can mitigate repetitive generation without significantly harming ICL, as these neurons primarily influence the output surface rather than the underlying pattern recognition.

While ablation confirms that late-segment repetition neurons are critical for ICL, their role extends beyond mere token replication. They exhibit high activation during pattern processing (average activation 1.23-1.49)~\footnote{Appendix~\ref{apx:ab_id}, Table~\ref{tab:apx_abhd_avgnr_l8} for Llama-3.1-8B} compared to near-zero or negative values in random neurons, indicating active involvement in ICL. We propose that they act as \textit{pattern executors}: final-stage components that translate abstract pattern signals, whether structural or semantic, into output decisions. Their position near the output layer enables a strong influence on logits, explaining the up to 60\% recall drop when ablated. This executor role is upstream-dependent: their activation drops significantly under induction-head ablation in structural tasks, supporting a detector-to-executor cascade. However, in semantic tasks, especially in Llama, their resilience suggests alternative pathways can sustain their activity. Thus, their function is not as independent agents, but as amplifiers of distributed pattern representations.


%
\section{Joint Ablation}
\label{sec:joint_ab}
\begin{figure}[t]
    \centering
    \includegraphics[width=\columnwidth]{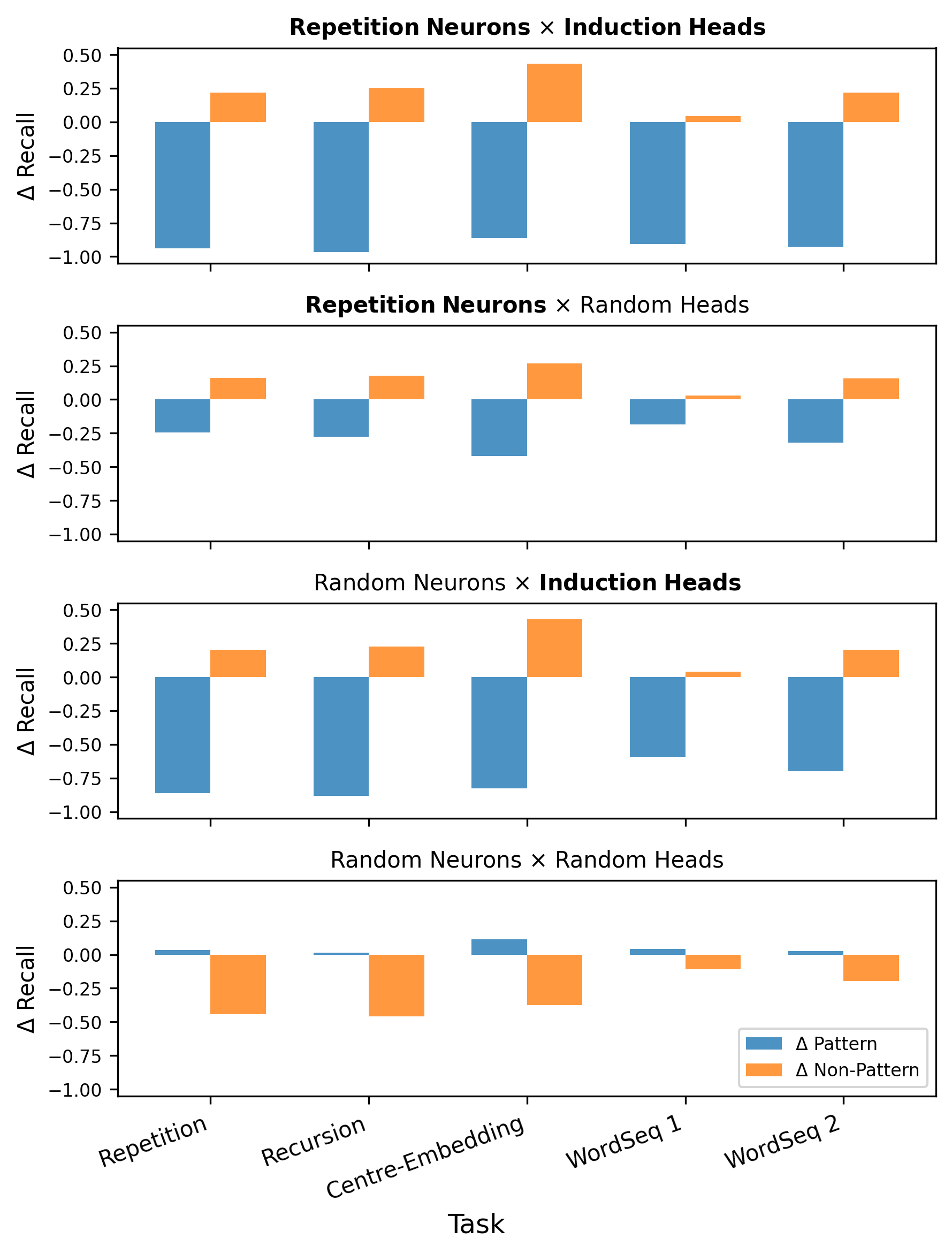}

    \caption{\(\Delta\)Recall for Pattern (blue) and Non-Pattern (orange) inputs on five tasks under four joint‐ablation regimes: (a) repetition neurons(top 250 from final segment [0.8–1.0]) \(\times\) induction heads (top 3\% by prefix-matching score), (b) repetition neurons \(\times\) random heads, (c) random neurons \(\times\) induction heads, and (d) random neurons \(\times\) random heads with 10-shot. The negative value means the performance drop after ablation.}
    
    \label{fig:dual_ablation}
\end{figure}

Experimental Results in \S\ref{sec:layer_wise} and \S\ref{sec:com_ab} indicate that both repetition neurons and induction heads play a central role in the model’s ability to capture pattern signals. 
Motivated by this, we perform a joint ablation of the top 3\% induction heads (by prefix‐matching score) and the top 250 repetition neurons (from the final 0.8–1.0 segment) to evaluate their combined effect.

Figure ~\ref{fig:dual_ablation} shows the impact of joint ablation of repetition neurons and induction heads.
Taking a pair including random ablation into account, the figure displays a total of four combinations. We observe a clear trend that ablating repetition neurons and induction heads damages the pattern subset significantly: three out of five tasks exhibit over 90\% recall degradation, reaching up to 96\% in certain settings~\footnote{Similar findings are confirmed on two additional models (see Appendix~\ref{apx:joint_ab}).}.
In contrast, when we randomly select both neurons and heads, the performance on the pattern subset is almost zero.
This observation underscores the critical role of both repetition neurons and induction heads in pattern detection.

The results also show that ablating induction heads generally causes a slightly larger or comparable drop in Pattern performance compared to ablating repetition neurons alone. This suggests that induction heads serve as primary detectors, while late-segment repetition neurons act as essential executors-both are indispensable, but operate at different stages of the cascade.” 
This demonstrates that the induction heads have the dominant role in handling repetitive inputs.
However, considering that the joint ablation caused the most damage, we conclude that the roles of the repetition neurons and induction heads are complementary.
As shown in Figures \ref{fig:ab_repr_layer} and \ref{fig:ab-id-class-llama8-10shot}, ablating repetition neurons alone in the 10-shot setting incurs a 40–68\% recall loss, and ablating induction heads alone causes a 50–77\% loss.  
The fact that joint ablation pushes this degradation beyond 90\% suggests a sequential, cascade mechanism, particularly in structural tasks, in which induction heads detect underlying pattern structure and subsequently activate late-segment repetition neurons to consolidate and execute the correct prediction. Disabling both components severs this cascade, resulting in a near-total collapse of the model’s pattern-recognition capability.
The experimental results in Figure \ref{fig:dual_ablation} on the Non-pattern subset also show a clear trend: the performance increases when ablating repetition neurons or induction heads, while it decreases when ablating both random components.
This result implies that removing critical components reduces the model's bias to detect the pattern structure, while preserving the capability of accepting non-pattern structure.

\section{Text Generation}
\label{sec:text_gen}
\begin{figure}
    \centering
    \includegraphics[width=1\linewidth]{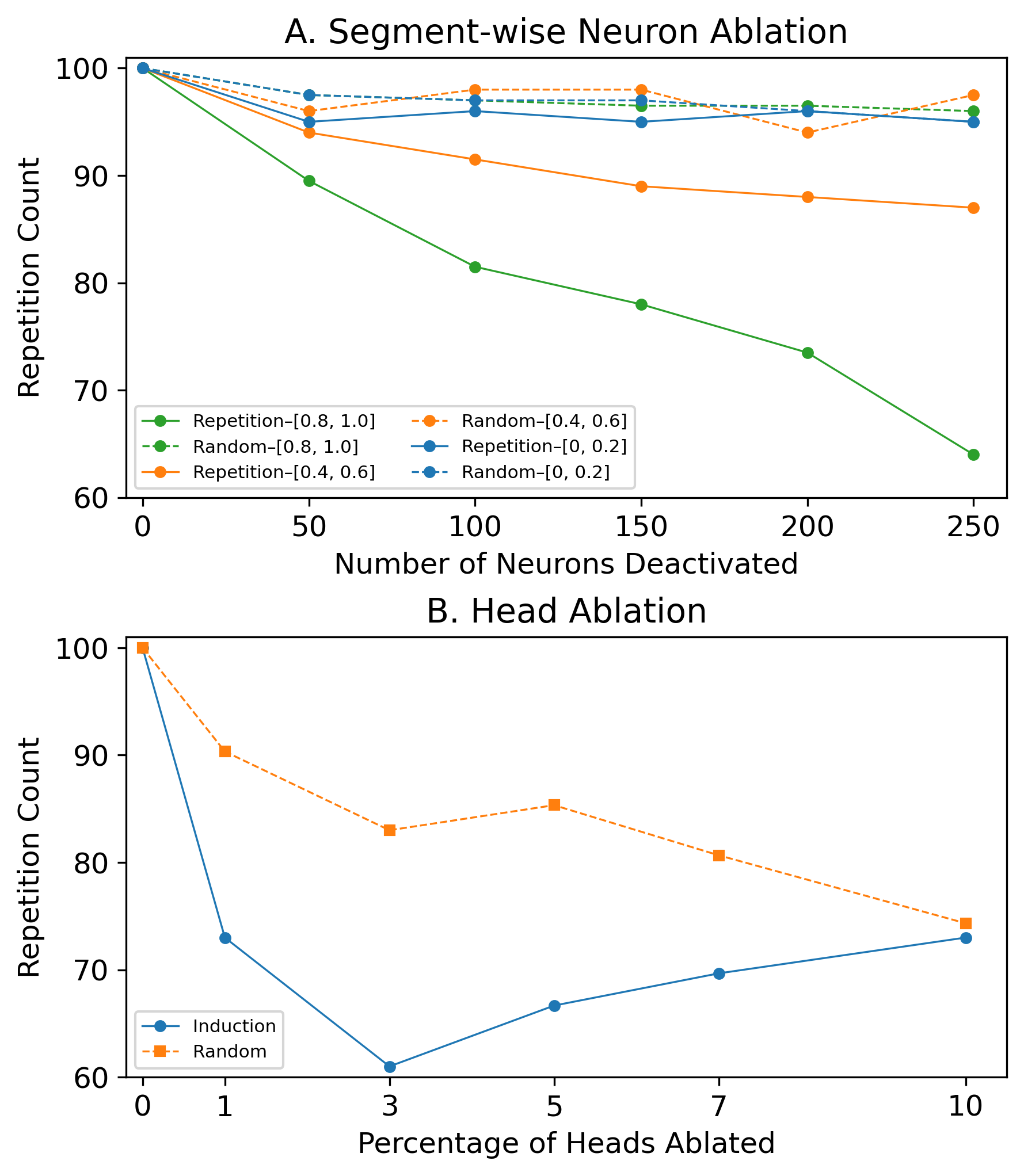}
    \caption{Effect of ablating repetition neurons and attention heads on generated repetition text. }
    \label{fig:textgen_nrhd}
\end{figure}

To investigate repetitive generation, distinct from prior sections that explored the roles and impacts of repetition neurons and induction heads in ICL tasks, we followed the approach of the prior work~\cite{hiraoka-inui-2025-repetition}. We prepared 100 greedily generated texts that originally include repetitive phrases.
Then we ablated the repetition neurons or induction heads at the onset of repetitive phrases during the generation process.

Figure~\ref{fig:textgen_nrhd} reveals that while ablating induction heads reduces repetition, random head ablation also has a noticeable effect. This indicates that repetitive outputs are influenced by a broader set of heads, making precise identification of repetition-specific heads challenging for fine-grained control.
 In addition, ablating the top 3 \% of induction heads causes a sharp drop in repetitive output, but removing larger fractions leads to a rebound in repetition. This pattern implies that only a small subset of heads is truly responsible, consistent with prior work on induction-head specificity. We observe a similar effect for repetition neurons in the final segment as ablating that group cuts repetition by up to 40 \%.
\begin{table*}[htbp!]
\centering
  \scriptsize
  \setlength{\tabcolsep}{2pt} 
  \resizebox{\linewidth}{!}{
  \begin{tabular}{@{}l cccc   ccc   ccc   ccc@{}}
    \toprule
    & & \multicolumn{3}{c}{BLEU} 
        & \multicolumn{3}{c}{ROUGE-L} 
        & \multicolumn{3}{c}{Distinct-2} 
        & \multicolumn{3}{c}{BERTScore (F1)} \\
    \cmidrule(lr){3-5} \cmidrule(lr){6-8} \cmidrule(lr){9-11} \cmidrule(lr){12-14}
    Segment        & Shot 
      & (1) & (2) & (3) 
      & (1) & (2) & (3) 
      & (1) & (2) & (3) 
      & (1) & (2) & (3) \\
    \midrule
    \( \lbrack 0.4,0.6 \rbrack \) & 0  & -0.804 & 0.185  & 0.047  & -0.186 & 0.080  & 0.300  & -0.004 & -0.003 & -0.005 & -0.003 & -0.001 & 0.000 \\%
    
    \( \lbrack 0.4,0.6 \rbrack \) & 5  & -0.622 & -0.164 & -0.199 & -0.552 & 0.008  & 0.182  & -0.005 & -0.003 & -0.003 & -0.001 & 0.000  & 0.000 \\%
    
    \( \lbrack 0.8,1.0 \rbrack \) & 0 & -12.048 & -6.747 & -5.810  
                 & -12.859 & -8.905 & -4.561  
                 & -0.010  & -0.060 & -0.070  
                 & -0.049  & -0.036 & -0.017 \\%
    
    \( \lbrack 0.8,1.0 \rbrack \) & 5 &  -4.033 & -3.552 & -8.249  
                 &  -2.409 & -3.915 & -7.532  
                 & -0.013  & -0.042 & -0.065  
                 & -0.004  & -0.006 & -0.013 \\%
    \bottomrule
  \end{tabular}
  }
  \caption{Change in LlaMA-3.1-8B performance (after–before) when ablating 250 repetition neurons in different segment layers on WMT19 (1050 samples: cs-en (1), de-en (2), zh–en (3)). Distinct-2 measures the ratio of unique bigrams to total bigrams in the generated translation, indicating the output’s lexical diversity.}
  \label{tab:wmt19-l3b}
\end{table*}

Regarding the repetitive output as a problematic behaviour of LLMs, these results straightforwardly suggest that ablating the induction heads or repetition neurons at the last segment mitigates this problem.
However, as we discussed in the above sections, ablating these components catastrophically degrades the ICL performance.
We argue it is counterproductive to damage the crucial ability of LLMs in order to reduce the repetitions.

To resolve this trade-off, we suggest fine control of the repetition ability of LLMs using segment-wise neuron ablation.
Specifically, ablating repetition neurons in the middle segment reduced the repetitive output while it did not significantly damage the ICL performance, as shown in Figure \ref{sec:layer_wise}.



\section{Downstream Tasks}
\label{sec:downstream}

We confirm our suggestion on ablating middle-layer repetition neurons to avoid unnecessary repetitions while preserving ICL performance by evaluating on two downstream tasks: sentiment analysis with binary classification and machine translation.
For the sentiment analysis, we used 1,000 samples of SST2 \citep{socher-etal-2013-recursive}.
The machine translation task used WMT19 \citep{ws-2019-machine-translation} with 1,050 samples, including cs-en, de-en, and zh-en.

Regarding SST2, we observe only small fluctuations in the F1-score of Llama-3.1-8B of repetition neurons of $[0.4,0.6]$~\footnote{We find the similar trends in both Qwen-2.5-7B and LlaMA-2-13B, see details in Appendix \ref{apx:text_gen}, Table \ref{tab:sst2-rep-ablation}.}, around a 0.04 point drop when ablating repetition neurons in the last segment layers, which is remarkable even compared with ablating random neurons. Similarly, as Table \ref{tab:wmt19-l3b} represents results of WMT19, middle-segment ablation reduces degenerate repetition with $\leq$1 point BLEU/ROUGE drop, whereas late-segment ablation collapses performance entirely. Additionally,  BERTScore-F1~\cite{zhangbertscore}, repetition neurons from the last segment decrease in 17/18 experiments, while the one from the middle segment decreases in just 3/18, and even increases slightly~\footnote{See detail results from Tables \ref{tab:sst2-rep-ablation} to \ref{tab:wmt19-last}, Appendix \ref{apx:text_gen}}.


\section{Conclusion}
We have presented a comprehensive analysis of how repetition neurons and induction heads causally underpin few-shot in-context learning. Our layer‐wise ablation experiments reveal a “two-peak” structure in repetition neurons - intermediate‐layer detectors and final‐layer replicators - where only the latter are critical for Pattern recognition, while ablating middle-segment neurons substantially reduces repetitive outputs and with minimal impact on ICL recall. In parallel, we demonstrate that a tiny subset of induction heads serves as primary pattern detectors: ablating just 1 \% causes modest performance drops, but 3 \% ablation collapses pattern recognition entirely. Finally, joint ablation confirms a sequential cascade in which induction heads activate repetition neurons to enforce repeated outputs.  

Crucially, the role of induction heads is not universal. While they are indispensable for structural patterns in Llama, their influence on late-segment repetition neurons weakens in semantic tasks - suggesting Llama employs alternative pathways for abstract pattern encoding. In contrast, Qwen-2.5-7B exhibits a tighter, more uniform coupling: ablating induction heads impairs ICL across all task types, indicating they serve as a general-purpose mechanism for both surface and semantic pattern recognition. This architectural divergence reveals that induction heads are not merely \enquote{copying circuits}, but can function as domain-general pattern encoders, depending on model design.

Our findings not only elucidate the internal mechanisms of pattern learning in LLMs but also suggest a practical intervention: selectively ablating middle‐segment repetition neurons offers a way to mitigate unwanted repetition while preserving in-context learning performance.

\textbf{Broader Applicability} While our main studies focus on the repetition problem, our three-step pipeline-(1) behavioral pattern detection, (2) neuron/head ranking, and (3) targeted ablation- can generalize well beyond repetition. For example, one could define a memorization metric (e.g., verbatim recovery of training snippets), rank \enquote{memorization neurons} via activation-difference or gradient saliency, and then causally probe them through segment-level ablations. This approach would systematically reveal and even steer the neural substrates of memorization, hallucination, or other emergent behaviors of group-specific skill neurons. 



\section*{Limitations}
Our study focuses specifically on understanding the causal role of repetition neurons and induction heads in LLMs within controlled, synthetic binary‐classification tasks designed to explicitly test repetitive pattern recognition in ICL. However, this setup entails several limitations:
\begin{itemize}
  \item \textbf{Intervention simplicity:} Our method of zeroing out activations of selected neurons and heads, while consistent with prior work, may oversimplify their nuanced functions; exploring graded activation scaling or neuron‐specific fine‐tuning could yield more precise control.
  \item \textbf{Functional attribution:} Characterizing repetition neurons as “detectors” or “replicators” rests solely on performance changes following ablation; complementary interpretability analyses, such as activation tracing or causal mediation, would further validate these functional attributions.
  \item \textbf{Ambiguous functional role of late-layer neurons:} While our ablation and activation analyses confirm that late-segment repetition neurons are indispensable for ICL, their precise computational role remains unclear. Their high activation and catastrophic impact upon removal suggest that they act as the final executors of pattern-based decisions, but whether they actively encode pattern semantics or merely amplify upstream signals remains unresolved. Future work could combine causal mediation, activation path tracing, or controlled probing to disentangle their role as pattern encoders versus output enforcers.
\end{itemize}


\section*{Acknowledgements}
This work was written independently, with minor phrasing assistance from ChatGPT. All interpretations and remaining errors are solely the authors’.
\bibliographystyle{acl_natbib}


\clearpage
\appendix

\section{Hardware}
All experiments were conducted on NVIDIA A40 GPU (48 GB VRAM).  
The data-generation process described in Appendix~\ref{apx:repnr} requires roughly 15–20 hours, depending on model size.  
Ablation experiments take about 5 minutes per setting.  
We used three random seeds (41, 42, and 43) to assess variability.

\section{In-context Learning dataset}
\label{apx:icl-dataset}
Sample for the ICL dataset of each task as in the Table \ref{tab:icl-sample}. For each shot setting, we ensure the uniform distribution of classes in the demonstration. 
\begin{table*}[ht]
  \centering
  \small
  \caption{Few-shot prompts and queries for each binary classification task}
  \label{tab:icl-sample}
  \begin{tabular}{@{} l l p{1.5cm} p{10cm} l @{}}
    \toprule
    \textbf{Task} & \textbf{Shots} & \textbf{Query} & \textbf{Prompt} & \textbf{GT} \\
    \midrule

    \multirow{2}{*}{Repetition}
      & 5  & \texttt{"U Z U Z"}  
        & Below is a list of various sequences. Your task is to classify each sequence. Use the examples provided to understand how to classify each sequence correctly. 
        Y K Y K: Foo \, I Y P X: Bar \, X K X K: Foo \, N L N L: Foo \, R S R S: Foo 
        U Z U Z:
        & Foo
    \\

      & 10 
        & \texttt{"U Z U Z"}  
        & Below is a list of various sequences. Your task is to classify each sequence. Use the examples provided to understand how to classify each sequence correctly. 
        N Y N Y: Foo \, N M Y C: Bar \, X Y X Y: Foo \, J Z X W: Bar \, X Q B V: Bar \, Z H Z H: Foo \, F J F J: Foo \, E Q E Q: Foo \, X K X K: Foo \, W U F Z: Bar 
        U Z U Z:
        & Foo
    \\[1ex]
    \hdashline
    \multirow{2}{*}{Recursion}
      & 5  & \texttt{"Q T S H"}  
        & Below is a list of various sequences. Your task is to classify each sequence. Use the examples provided to understand how to classify each sequence correctly. 
        Z W W W: Foo \, W J J J: Foo \, N T T T: Foo \, X H B I: Bar \, I M H W: Bar 
        Q T S H:
        & Bar
    \\
    
      & 10 & \texttt{"Q T S H"}  
        & Below is a list of various sequences. Your task is to classify each sequence. Use the examples provided to understand how to classify each sequence correctly. 
        Y Y Q F: Bar \, N P P P: Foo \, J Q K H: Bar \, N Q Q Q: Foo \, M Q Q Q: Foo \, K O O O: Foo \, I Y Y Y: Foo \, G A R S: Bar \, T O T Q: Bar \, A S S S: Foo 
        Q T S H:
        & Bar
    \\[1ex]
    \hdashline
    \multirow{2}{*}{Centre-embedding}
      & 5  & \texttt{"V A S A V"}  
        & Below is a list of various sequences. Your task is to classify each sequence. Use the examples provided to understand how to classify each sequence correctly. 
        M L Q L M: Foo \, K O N O K: Foo \, W O E O W: Foo \, Z U J U Z: Foo \, B C Z X C: Bar 
        V A S A V:
        & Foo
    \\

      & 10 & \texttt{"V A S A V"}  
        & Below is a list of various sequences. Your task is to classify each sequence. Use the examples provided to understand how to classify each sequence correctly. 
        X C K C X: Foo \, I N A N I: Foo \, K H M H K: Foo \, S X G F W: Bar \, E U D Z W: Bar \, K H Z L N: Bar \, D G Z K I: Bar \, A M L M A: Foo \, V O S Z T: Bar \, P T E A O: Bar 
        V A S A V:
        & Foo
    \\[1ex]
    \hdashline
    \multirow{2}{*}{WordSeq1}
      & 5  & \texttt{"mango lion"}  
        & Below is a list of various sequences. Your task is to classify each sequence. Use the examples provided to understand how to classify each sequence correctly. 
        papaya cat: Foo \, knee helicopter: Bar \, carrot drum: Bar \, kiwi shark: Foo \, apple rabbit: Foo 
        mango lion:
        & Foo
    \\

      & 10 & \texttt{"mango lion"}  
        & Below is a list of various sequences. Your task is to classify each sequence. Use the examples provided to understand how to classify each sequence correctly. 
        nectarine horse: Foo \, knee car: Bar \, broccoli skateboard: Bar \, ankle helicopter: Bar \, papaya fox: Foo \, fig horse: Foo \, leg boat: Bar \, toe accordion: Bar \, cherry fox: Foo \, lettuce skateboard: Bar 
        mango lion:
        & Foo
    \\[1ex]
    \hdashline
    \multirow{2}{*}{WordSeq2}
      & 5  & \texttt{"cucumber plane"}  
        & Below is a list of various sequences. Your task is to classify each sequence. Use the examples provided to understand how to classify each sequence correctly. 
        lemon dog: Bar \, beet bicycle: Foo \, strawberry doctor: Bar \, bookcase whale: Bar \, pea helicopter: Foo 
        cucumber plane:
        & Foo
    \\

      & 10 & \texttt{"cucumber plane"}  
        & Below is a list of various sequences. Your task is to classify each sequence. Use the examples provided to understand how to classify each sequence correctly. 
        onion yacht: Foo \, pea scooter: Foo \, pea skateboard: Foo \, celery skateboard: Foo \, radish minivan: Foo \, grape giraffe: Bar \, potato helicopter: Foo \, couch teacher: Bar \, tomato boat: Foo \, bed whale: Bar 
        cucumber plane:
        & Foo
    \\

    \bottomrule
  \end{tabular}
\end{table*}
\section{Repetition Neurons Identification} 
\label{apx:repnr}

\begin{figure}[ht]
    \centering
    \begin{subfigure}[t]{0.5\textwidth}
        \centering
        \includegraphics[width=\linewidth]{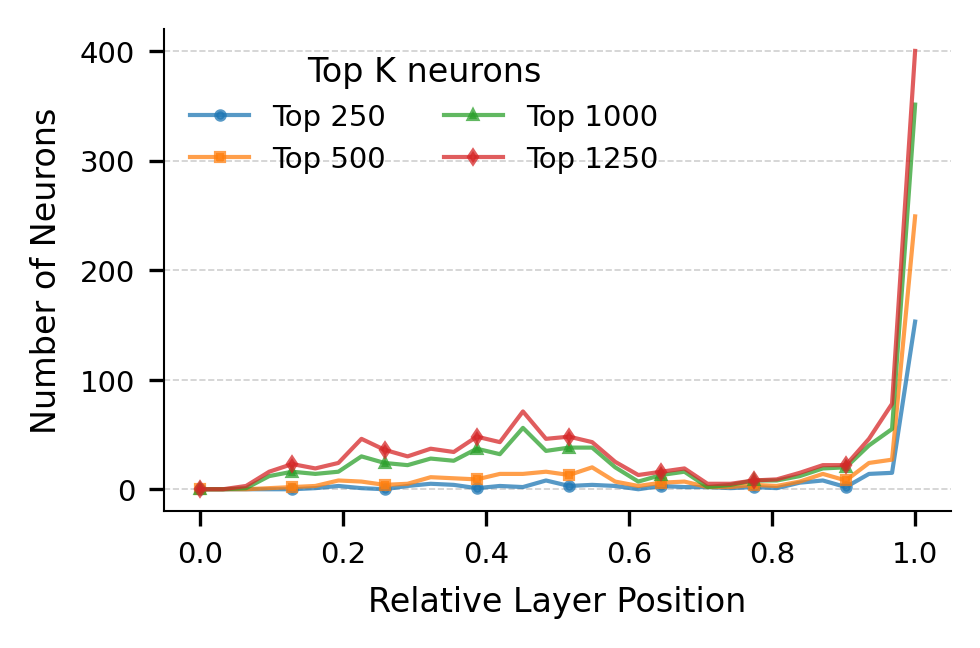}
        \caption{LLaMA-3.1-8B}
        \label{fig:rep-dist-8B}
    \end{subfigure}
    \\[1em]
    \begin{subfigure}[t]{0.5\textwidth}
        \centering
        \includegraphics[width=\linewidth]{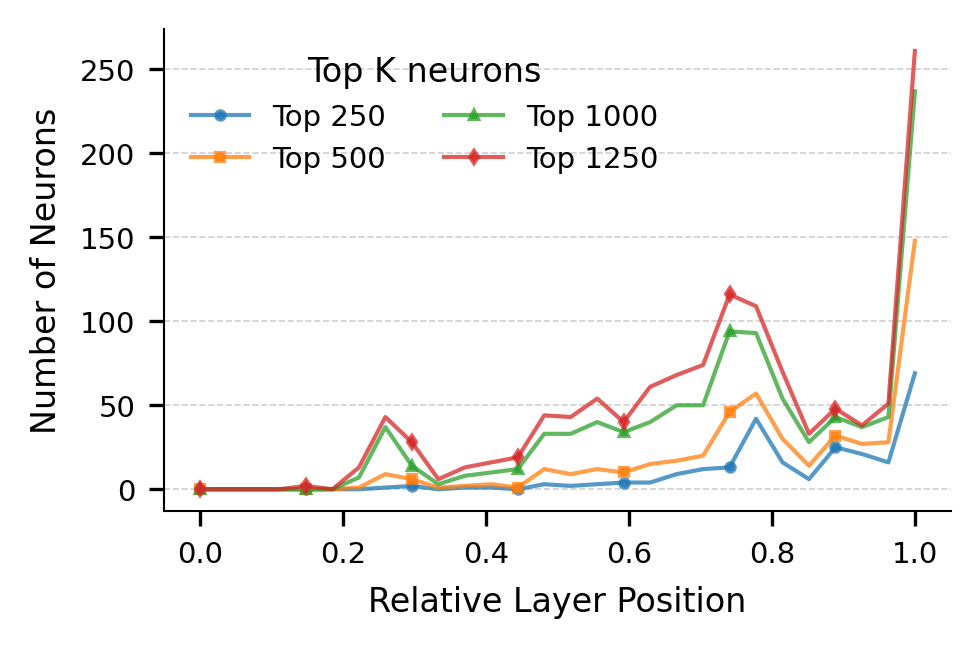}
        \caption{Qwen-2.5-7B}
        \label{fig:rep-dist-qwen}
    \end{subfigure}
    \\[1em]
    \begin{subfigure}[t]{0.5\textwidth}
        \centering
        \includegraphics[width=\linewidth]{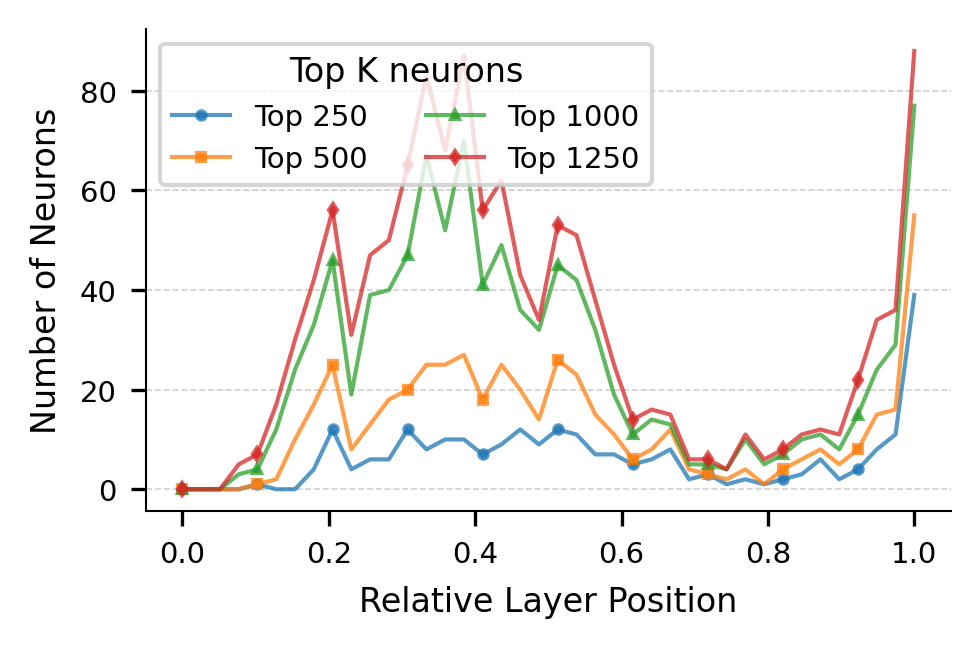}
        \caption{LLaMA-2-13B}
        \label{fig:rep-dist-13B}
    \end{subfigure}
    \caption{Layer-wise distribution of repetition neurons across three models. For each top-$K$ threshold (250, 500, 1000, 1250) ranked by $\Delta_n$, we count how many repetition neurons fall into each relative layer position (0 = first layer; 1 = last layer). The y-axis shows the number of repetition neurons per layer.}
    \label{fig:rep-dist}
\end{figure}

To generate repetitive text following the methodology outlined by \citet{hiraoka-inui-2025-repetition}, we begin by sampling the first 10 tokens using the \verb|generate()| function with \verb|temperature = 1.0|. The remainder of the sequence is generated using greedy decoding, with a maximum length of 200 tokens. A sample is classified as repetitive if it contains a 10-gram token sequence that repeats three times at regular intervals within a span of 100 tokens. We also applied filters to remove any texts where fewer than 50 tokens appeared before the second occurrence of the repeated pattern or fewer than 50 tokens followed the last repetition. The distribution of repetition neurons is as in Figure \ref{fig:rep-dist}.

\section{Layer-wise Repetition Neuron Ablation}
\label{apx:lw_repnr}

\begin{figure*}
    \centering
    \includegraphics[width=1\linewidth]{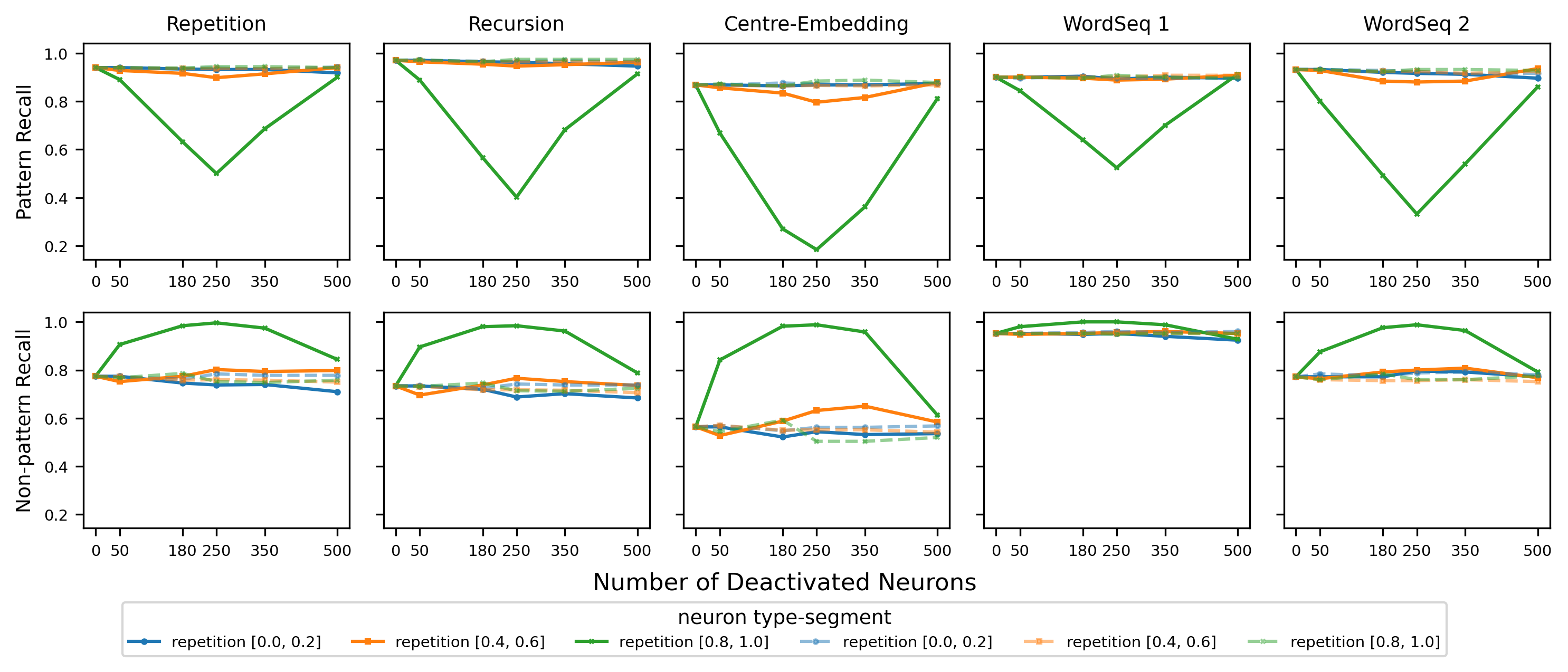}
    \caption{LlaMA-3.1-8B performance after ablating layer-wise neurons on two classes in 10-shot setting}
    \label{fig:ab-nr-500-l38b}
\end{figure*}

\begin{figure*}
    \centering
    \includegraphics[width=1\linewidth]{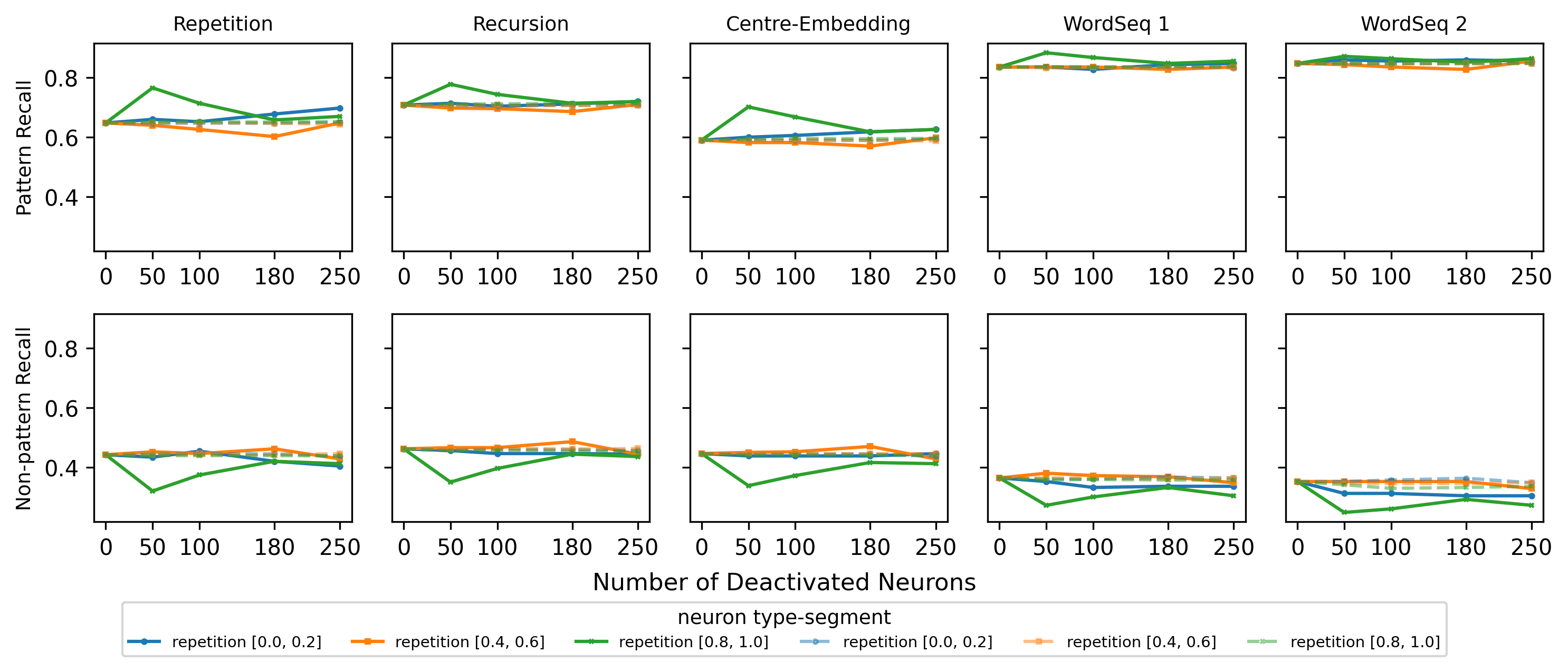}
    \caption{LlaMA-2-13B performance after ablating layer-wise neurons on two classes in 5-shot setting}
    \label{fig:ab-nr-l213b}
\end{figure*}

We report recall on two classes after intervening in three models with 5-shot and 10-shot settings from Figure \ref{fig:ab-nr-500-l38b} to Figure \ref{fig:ab-nr-l213b}.

\begin{figure*}
    \centering
    \includegraphics[width=1\linewidth]{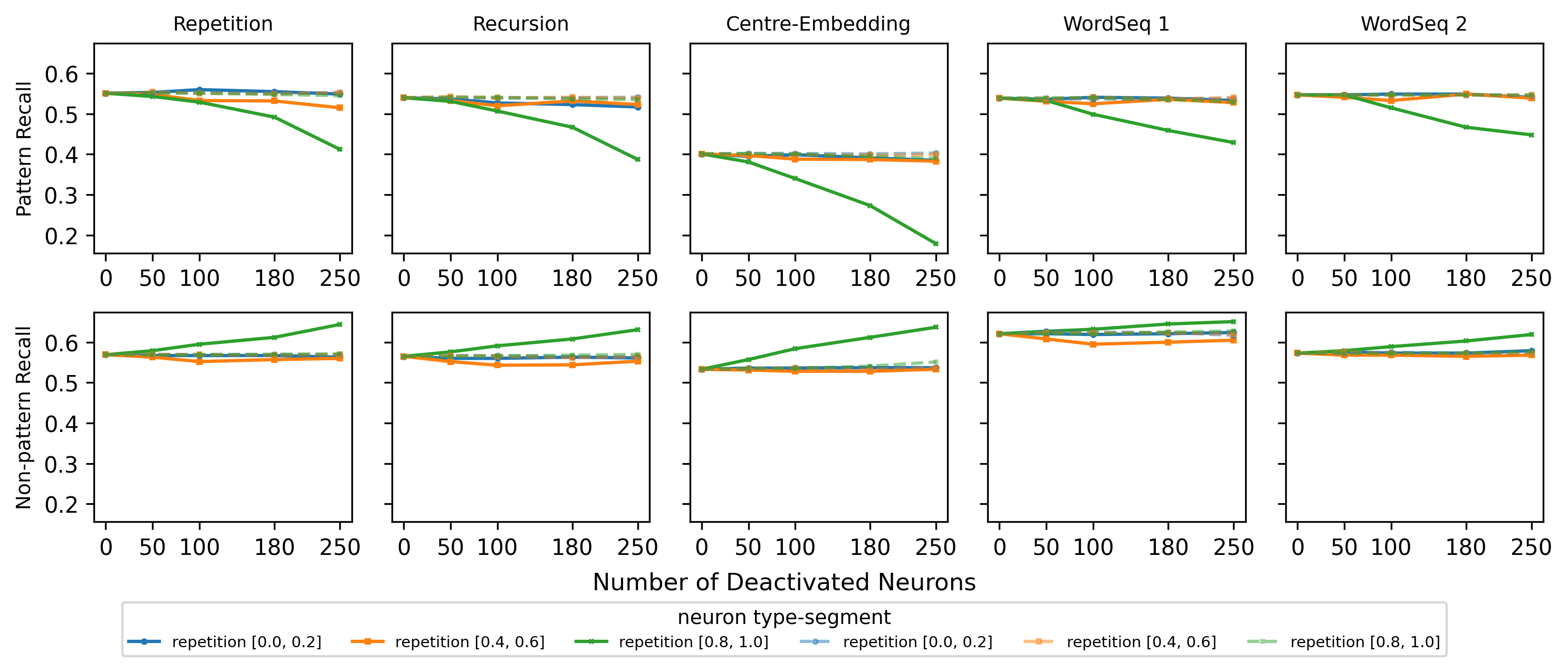}
    \caption{Qwen2.5-7B performance after ablating layer-wise neurons on two classes in 10-shot setting}
    \label{fig:ab-id-class-llama8}
\end{figure*}

\begin{figure*}
    \centering
    \includegraphics[width=1\linewidth]{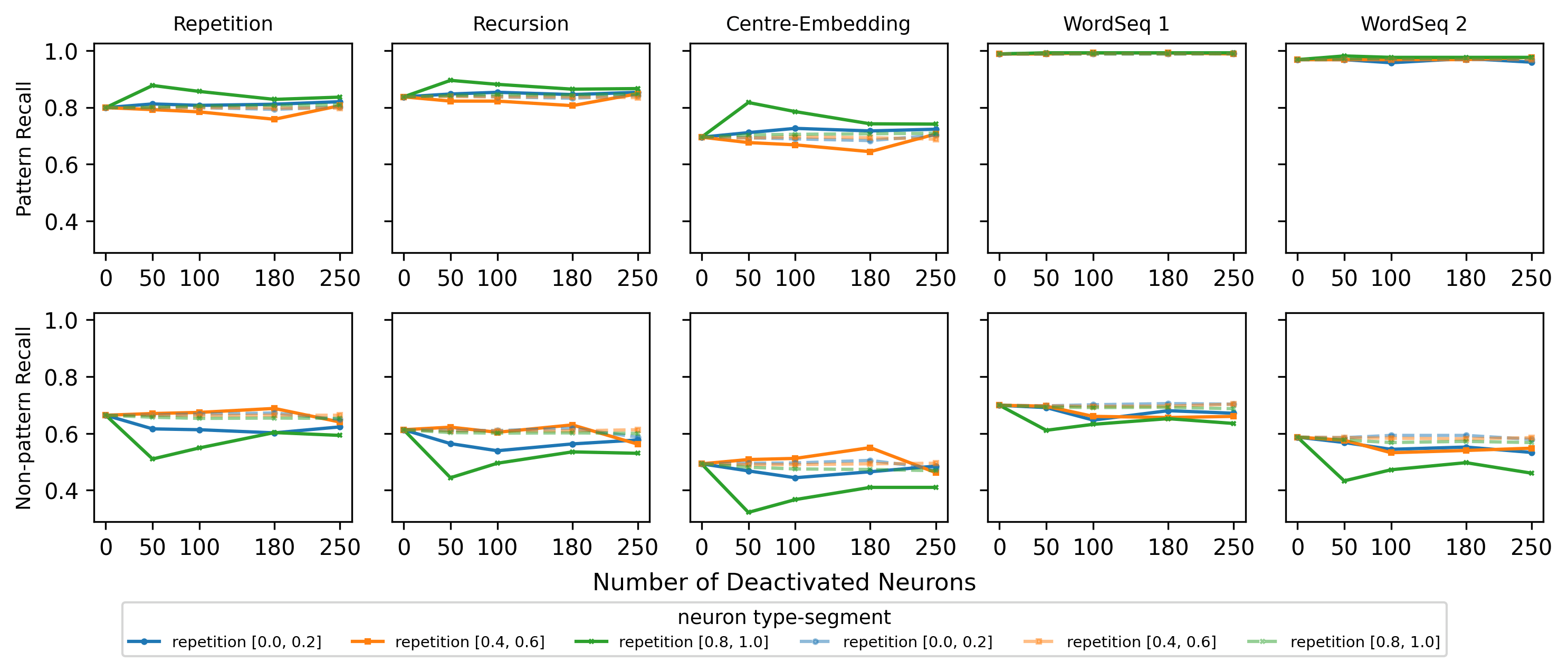}
    \caption{LlaMA-2-13B performance after ablating layer-wise neurons on two classes in 10-shot setting}
    \label{fig:ab-id-class-llama8}
\end{figure*}

\begin{figure*}
    \centering
    \includegraphics[width=1\linewidth]{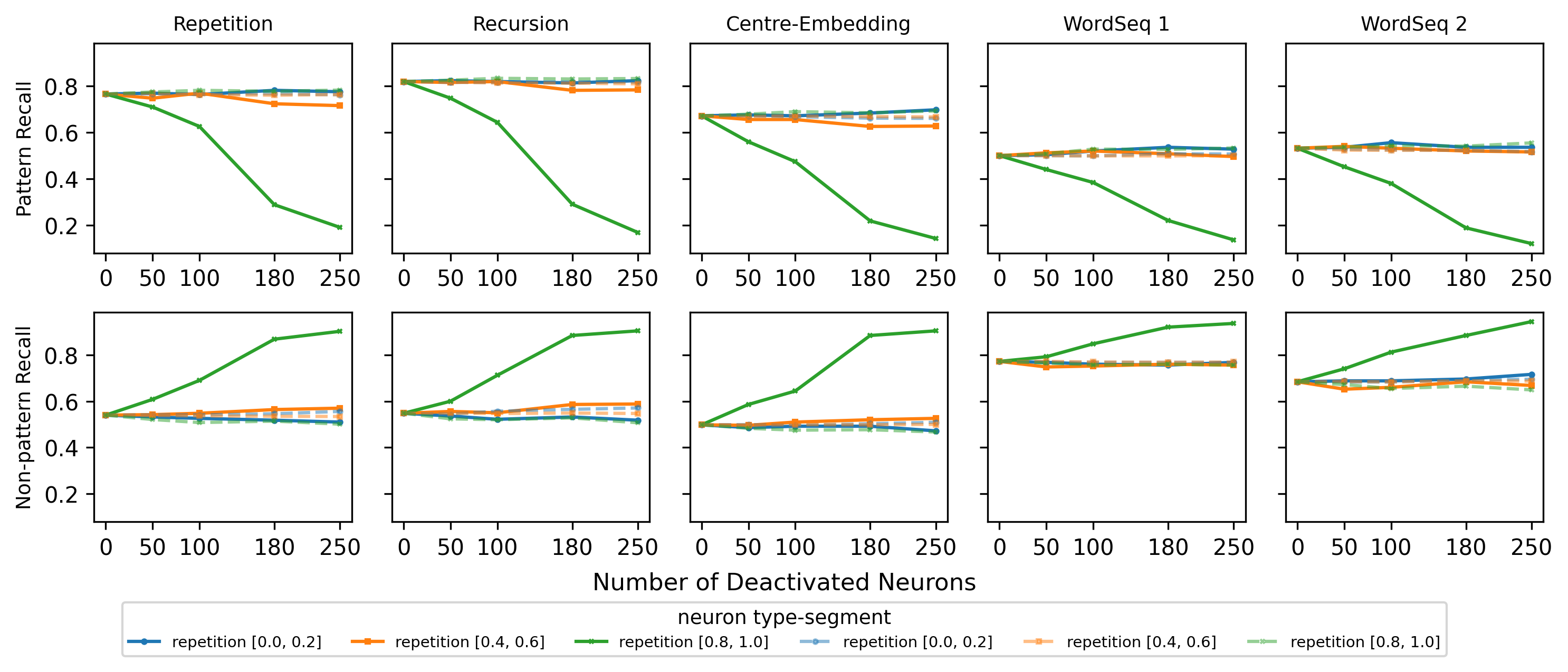}
    \caption{LlaMA-3.1-8B performance after ablating layer-wise neurons on two classes in 5-shot setting}
    \label{fig:ab-id-class-llama8}
\end{figure*}

\begin{figure*}
    \centering
    \includegraphics[width=1\linewidth]{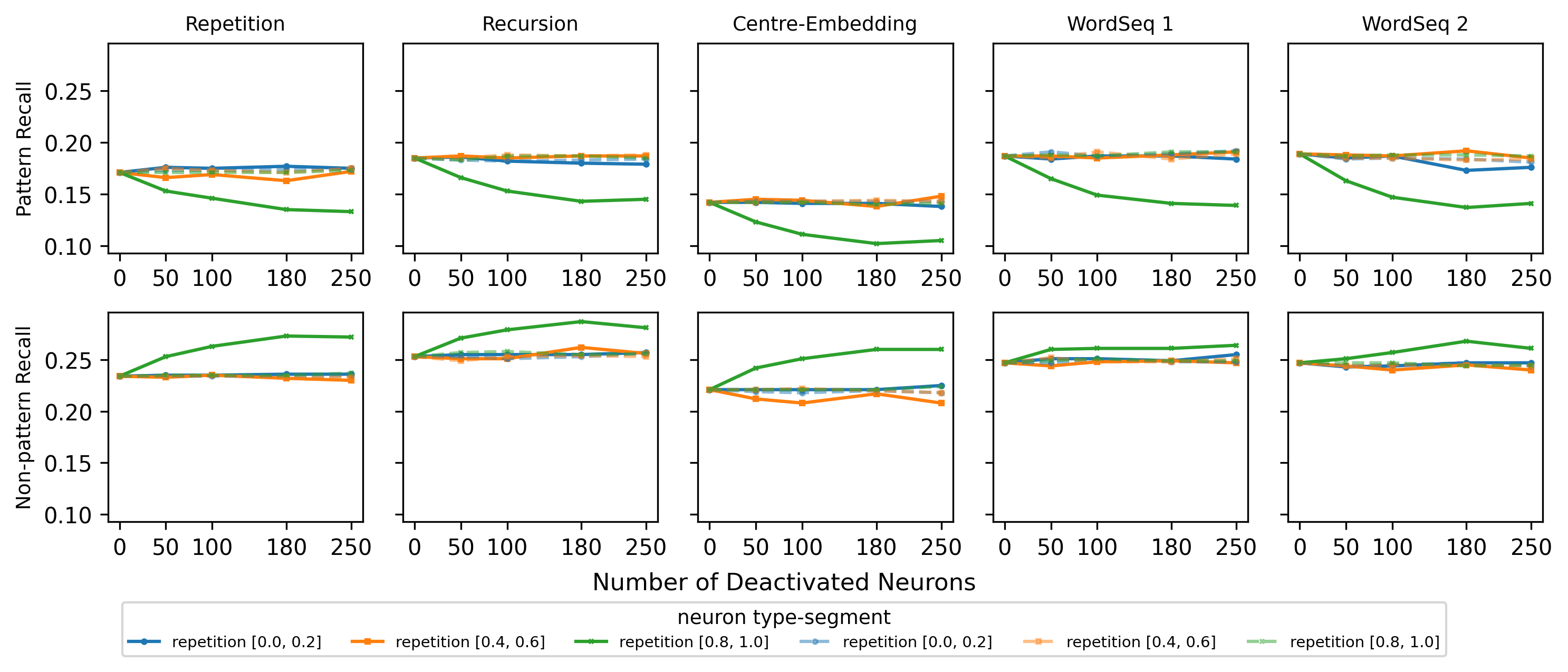}
    \caption{Qwen2.5-7B performance after ablating layer-wise neurons on two classes in 5-shot setting}
    \label{fig:ab-id-class-llama8}
\end{figure*}

\section{Ablation Induction Heads}
\label{apx:ab_id}

\begin{table*}[t]
\centering
\small

\setlength{\tabcolsep}{3pt}
\begin{tabular}{l l
   *{4}{c}
   *{4}{c}}
\toprule
\multirow{2}{*}{Task} & \multirow{2}{*}{Mode}
  & \multicolumn{4}{c}{Pattern}
  & \multicolumn{4}{c}{Non-Pattern} \\
\cmidrule(lr){3-6} \cmidrule(lr){7-10}
  & 
  & \multicolumn{2}{c}{5-shot} & \multicolumn{2}{c}{10-shot}
  & \multicolumn{2}{c}{5-shot} & \multicolumn{2}{c}{10-shot} \\
\cmidrule(lr){3-4} \cmidrule(lr){5-6} \cmidrule(lr){7-8} \cmidrule(lr){9-10}
  & 
  & 1\,\% & 3\,\% & 1\,\% & 3\,\% 
  & 1\,\% & 3\,\% & 1\,\% & 3\,\% \\
\midrule
\multirow{2}{*}{Repetition}
  & ind   & –0.021 & –0.643 & –0.025 & –0.767
          & –0.171 &  0.367 & –0.444 &  0.145 \\
  & ran   & –0.039 &  0.055 & –0.011 & –0.025
          &  0.009 & –0.075 & –0.041 & –0.049 \\
\addlinespace
\multirow{2}{*}{Recursion}
  & ind   & –0.055 & –0.721 &  0.001 & –0.771
          & –0.111 &  0.371 & –0.465 &  0.187 \\
  & ran   & –0.072 &  0.032 & –0.002 & –0.029
          &  0.007 & –0.072 & –0.044 & –0.060 \\
\addlinespace
\multirow{2}{*}{Centre-Embedding}
  & ind   &  0.019 & –0.577 &  0.052 & –0.754
          & –0.115 &  0.429 & –0.381 &  0.379 \\
  & ran   & –0.045 &  0.061 &  0.009 & –0.002
          &  0.025 & –0.098 & –0.017 & –0.021 \\
\addlinespace
\multirow{2}{*}{WordSeq 1}
  & ind   &  0.015 & –0.417 & –0.008 & –0.531
          & –0.139 &  0.189 & –0.188 &  0.032 \\
  & ran   & –0.017 &  0.089 & –0.004 & –0.025
          &  0.025 & –0.059 & –0.004 & –0.017 \\
\addlinespace
\multirow{2}{*}{WordSeq 2}
  & ind   & –0.004 & –0.453 & –0.012 & –0.632
          & –0.115 &  0.265 & –0.229 &  0.183 \\
  & ran   & –0.025 &  0.073 & –0.021 & –0.051
          &  0.028 & –0.065 & –0.017 &  0.009 \\
\bottomrule
\end{tabular}
\caption{LlaMA-3.1-8B $\Delta$Recall at 1\,\% and 3\,\% head ablation relative to baseline, for 5-shot vs 10-shot.}
\label{tab:hd_delta}
\end{table*}

Table \ref{tab:hd_delta} shows the change in the recall when ablating the top 1\% and 3\% induction heads of Llama-3.1-8B. While there is a decreasing trend in Pattern class for both cases, Non-Pattern exhibits nuances. We argue the case of Non-Pattern depends on model types as in Qwen model, we observe reverse trend to Llama. For Pattern, induction heads have a causal relationship as ablating them reduces model performance.
\begin{figure*}
    \centering
    \includegraphics[width=1\linewidth]{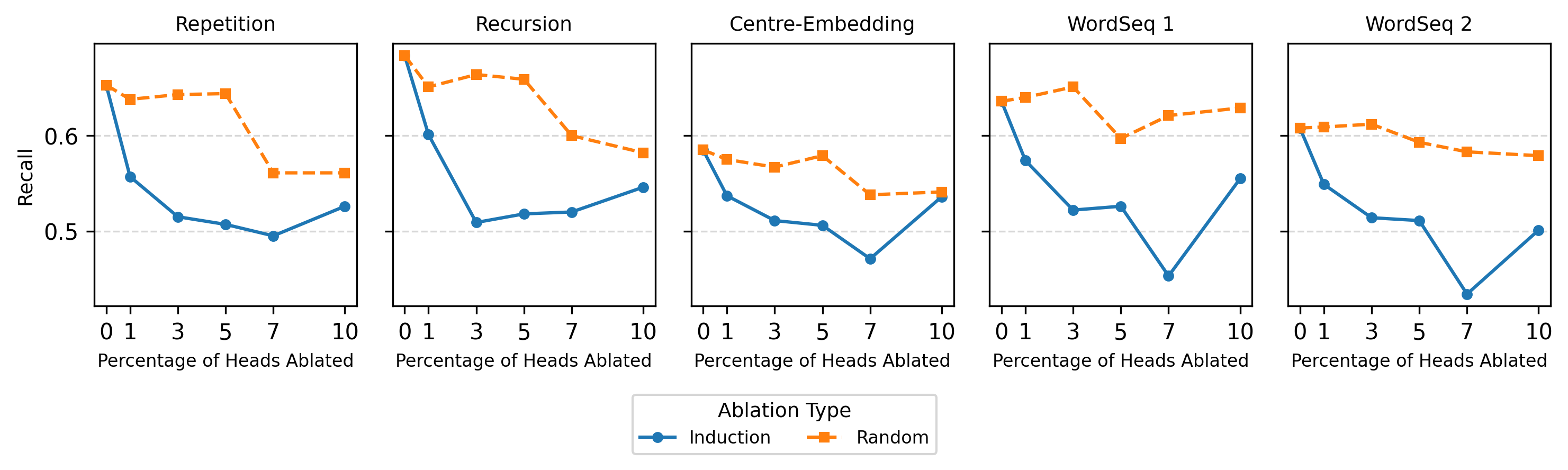}
    \caption{Llama-3.1-8B performance after ablating induction heads with top highest prefix scores in 5-shot setting}
    \label{fig:ab-id-llama8}
\end{figure*}

\begin{figure*}
    \centering
    \includegraphics[width=1\linewidth]{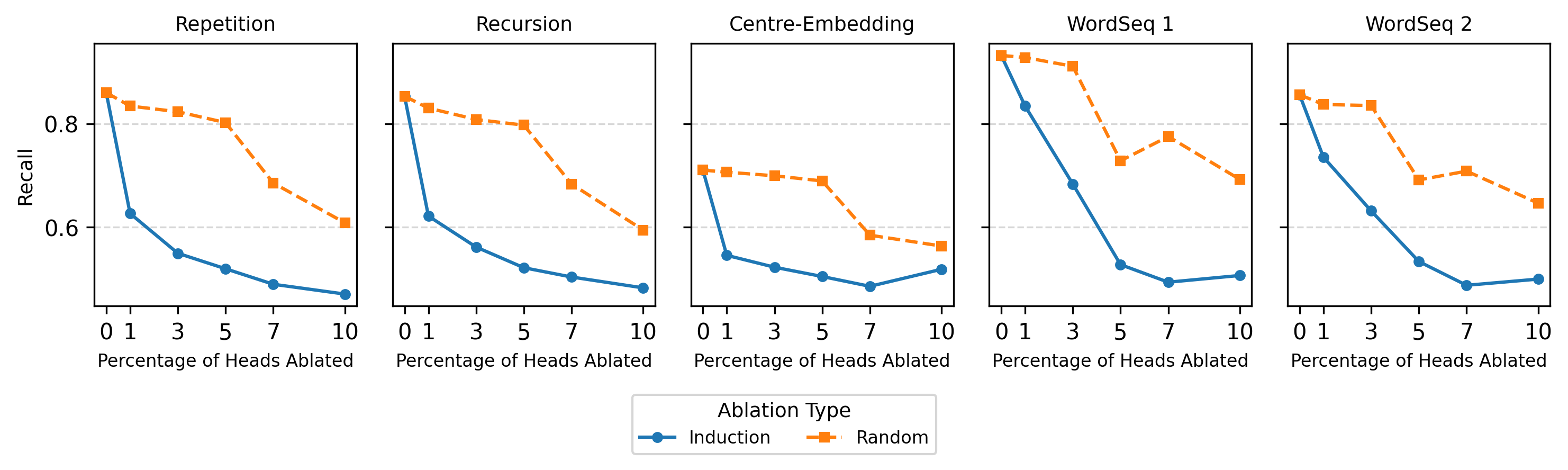}
    \caption{LlaMA-3.1-8B performance after ablating induction heads with top highest prefix scores in 10-shot setting}
    \label{fig:ab-id-llama8}
\end{figure*}

\begin{figure*}
    \centering
    \includegraphics[width=1\linewidth]{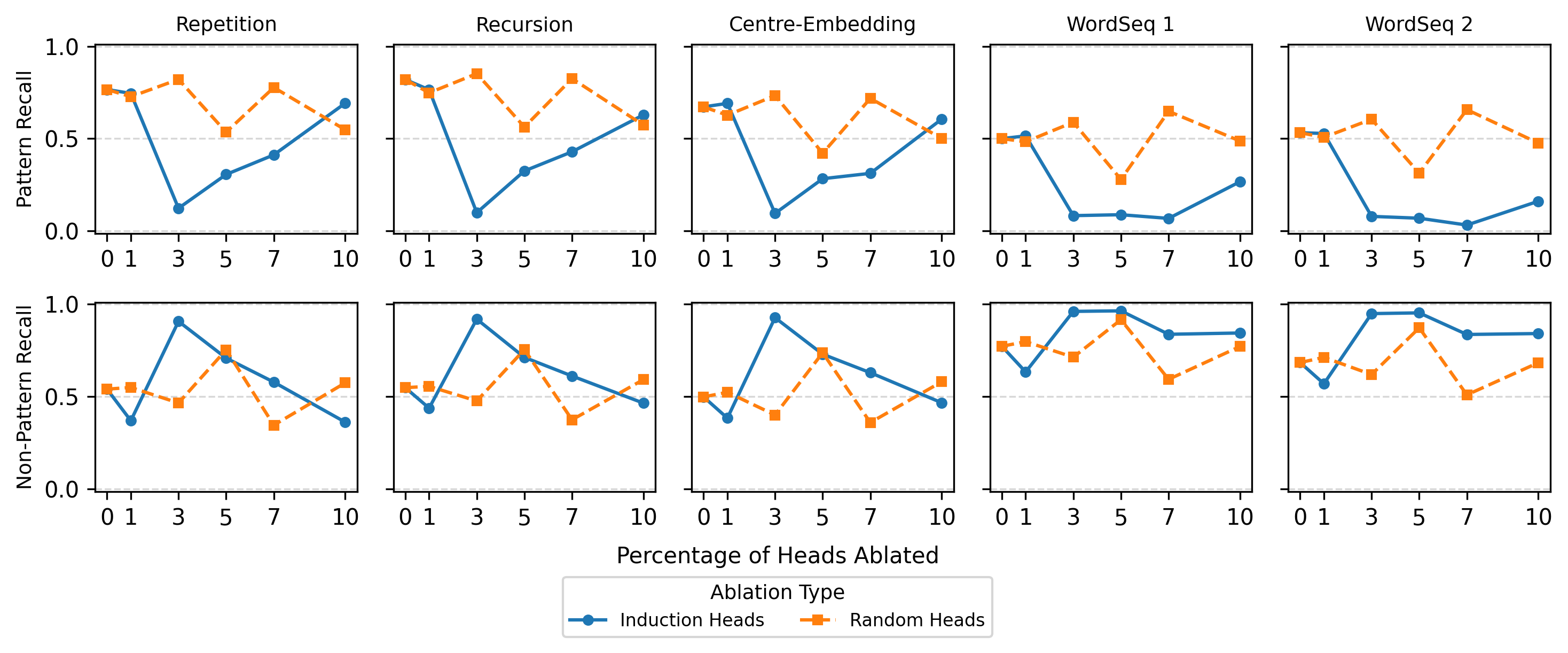}
    \caption{LlaMA-3.1-8B performance after ablating induction heads with the top highest prefix scores on two classes in the 5-shot setting.}
    \label{fig:ab-id-class-llama8-5shot}
\end{figure*}

\begin{figure*}
    \centering
    \includegraphics[width=1\linewidth]{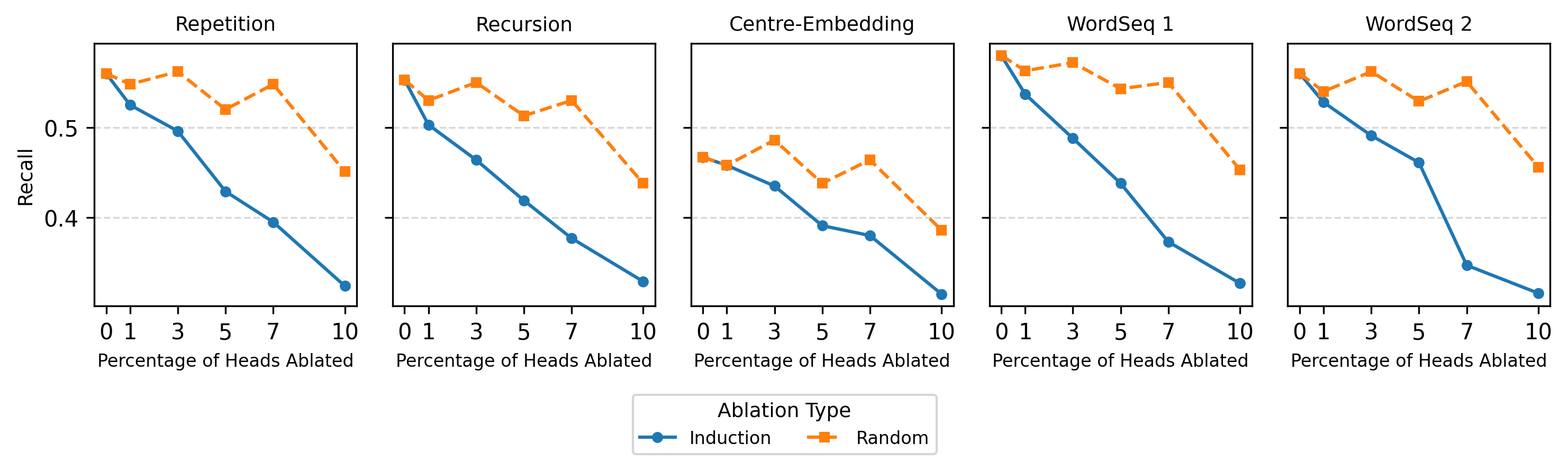}
    \caption{Qwen2.5-7B performance after ablating induction heads with top highest prefix scores in 10-shot setting}
    \label{fig:ab-id-llama8}
\end{figure*}
\begin{figure*}
    \centering
    \includegraphics[width=1\linewidth]{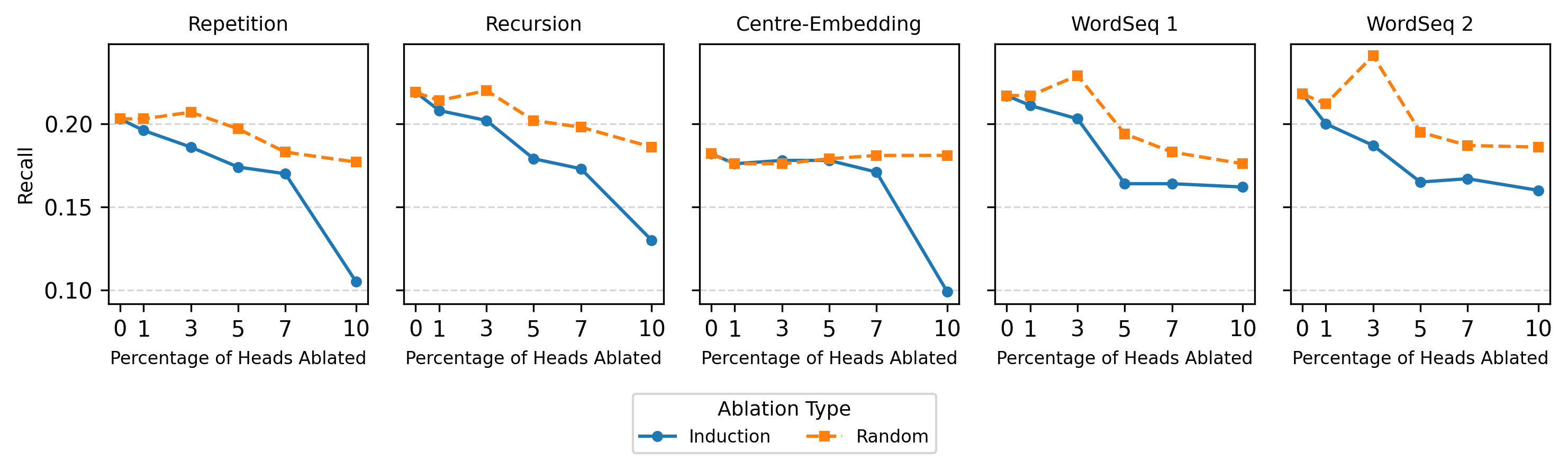}
    \caption{Qwen2.5-7B performance after ablating induction heads with top highest prefix scores in 5-shot setting}
    \label{fig:ab-id-llama8}
\end{figure*}
\begin{figure*}
    \centering
    \includegraphics[width=1\linewidth]{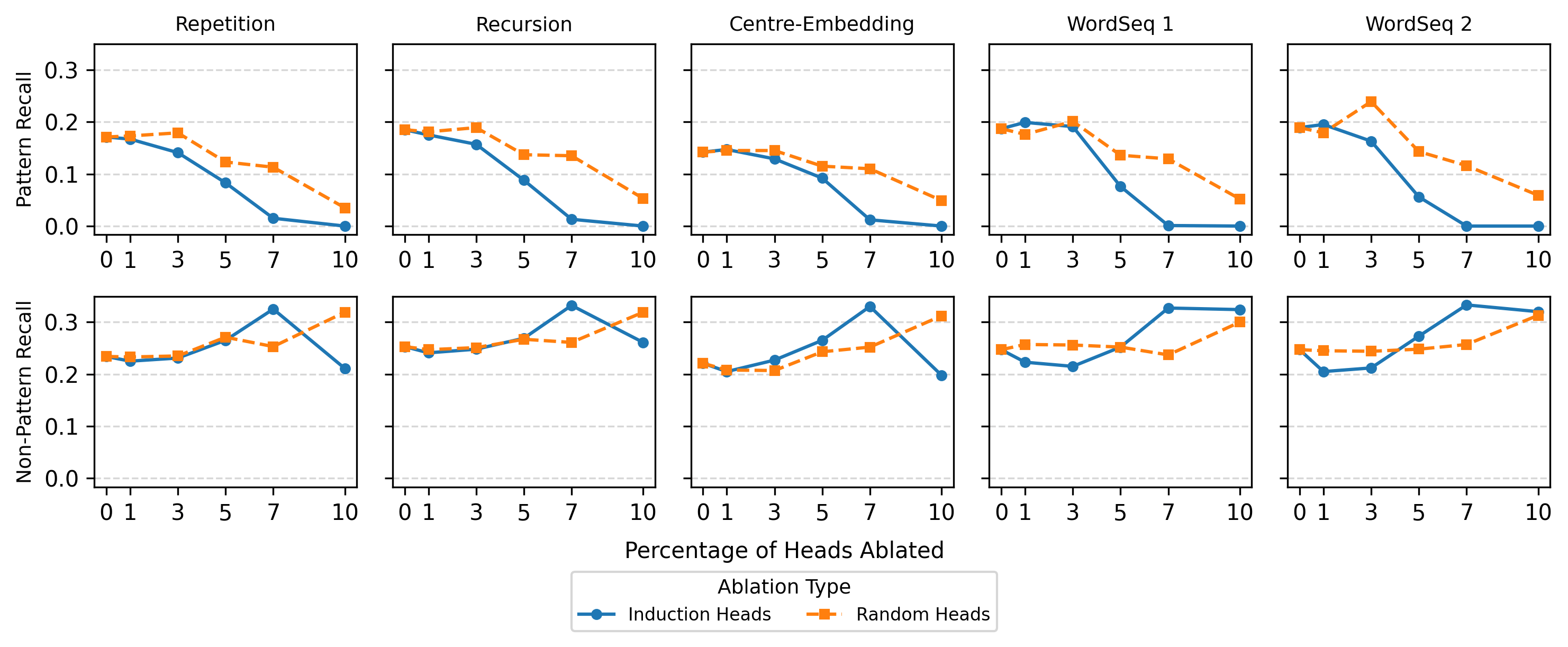}
    \caption{Qwen2.5-7B performance after ablating induction heads with top highest prefix scores on two classes in 5-shot setting}
\end{figure*}

\begin{figure*}
    \centering
    \includegraphics[width=1\linewidth]{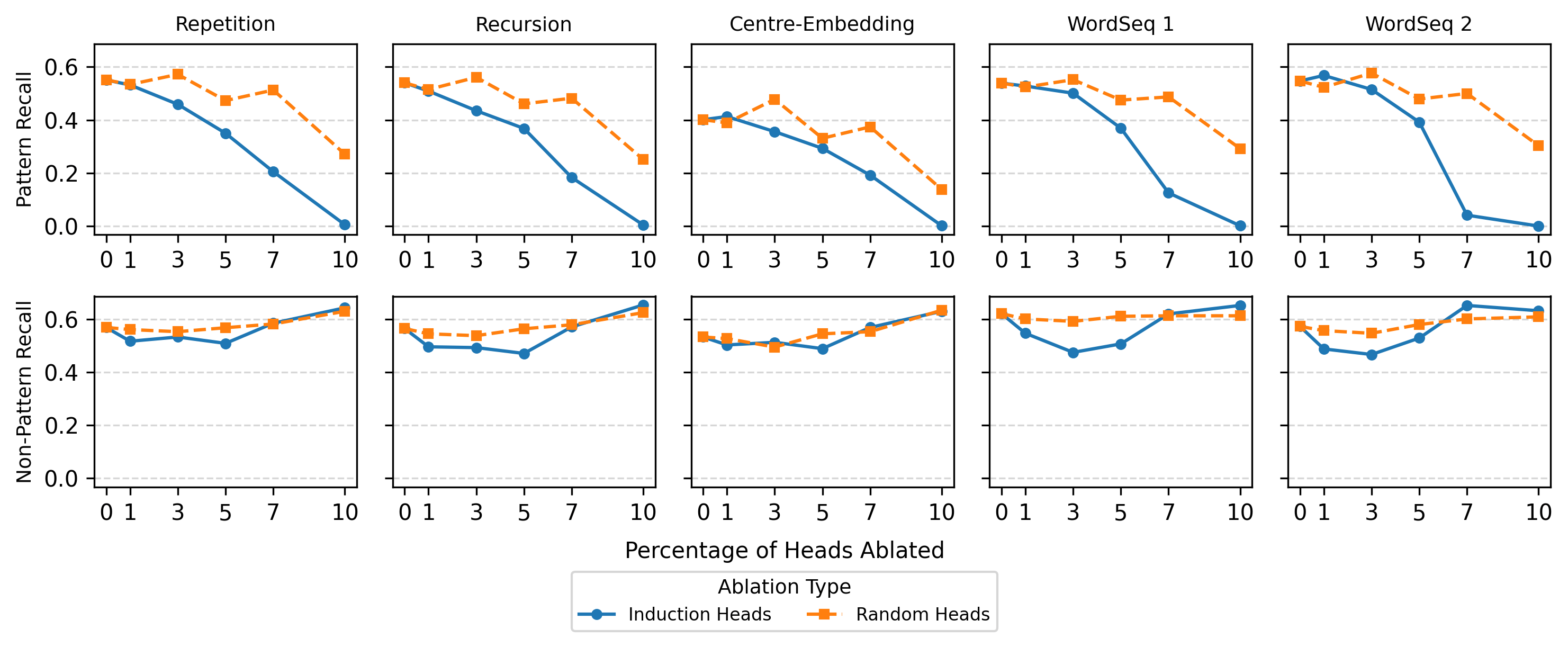}
    \caption{Qwen2.5-7B performance after ablating induction heads with top highest prefix scores on two classes in 10-shot setting}
    \label{fig:ab-id-class-qwen-10shot}
\end{figure*}

\begin{figure*}
    \centering
    \includegraphics[width=1\linewidth]{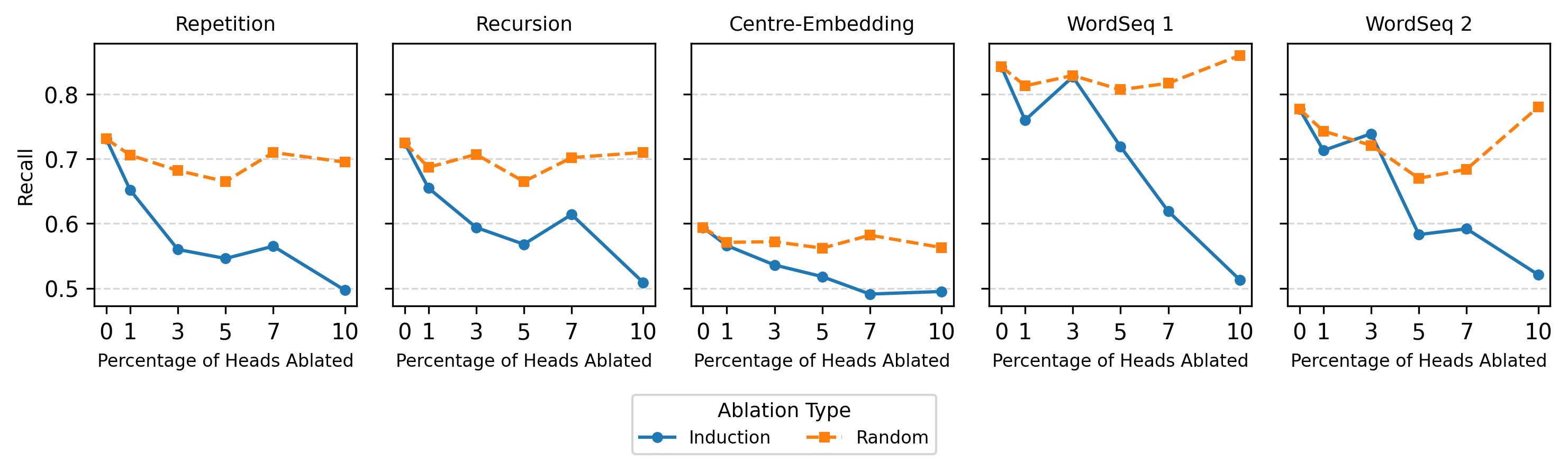}
    \caption{LlaMA-2-13B performance after ablating induction heads with top highest prefix scores in 10-shot setting}
    \label{fig:ab-id-llama8}
\end{figure*}

\begin{figure*}
    \centering
    \includegraphics[width=1\linewidth]{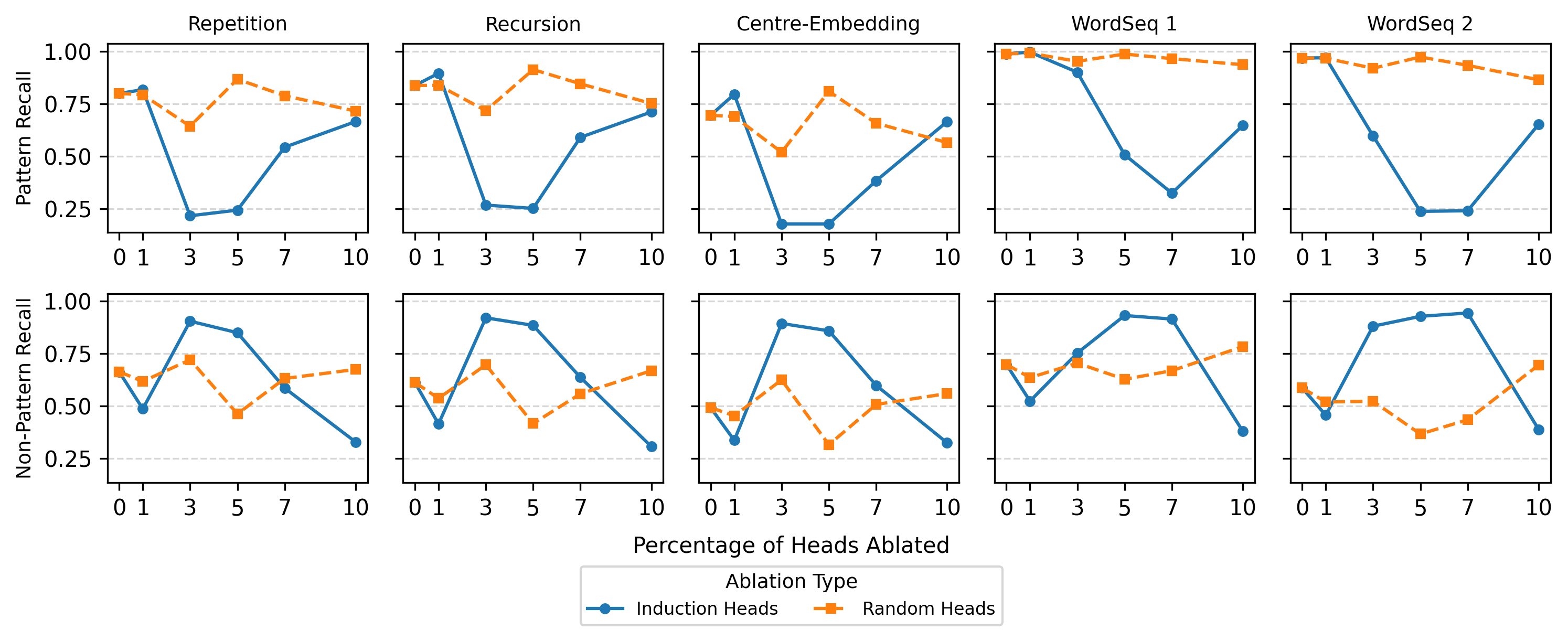}
    \caption{LlaMA-2-13B performance after ablating induction heads with top highest prefix scores on two classes in 10-shot setting}
    \label{fig:ab-id-class-llama13-10shot}
\end{figure*}

We also show the average activation value of repetition neurons and random neurons in the same segments when ablating induction heads or random heads from Table \ref{tab:apx_abhd_avgnr_l8} to Table \ref{tab:apx_abhd_avgnr_l13}. One consistent trend is that the value of repetition neurons is always higher than that of random neurons under the same conditions, which indicates their stronger activation for ICL tasks. The gap between the two groups of neurons increases with the depth of the layers. For example, while the average activation of random neurons in Qwen-2.5-7B is around -0.05, the repetition neurons are recorded with a value range from 3.8 to 4.3, which are significant gaps.
\begin{table*}[t]
\centering
\footnotesize
\setlength{\tabcolsep}{4.8pt}
\begin{tabular}{llrrrrrrrrrr}
\toprule
\multirow{2}{*}{\textbf{Segment}} & \multirow{2}{*}{\textbf{Cond.}} &
\multicolumn{2}{c}{\textbf{Repetition}} &
\multicolumn{2}{c}{\textbf{Recursion}} &
\multicolumn{2}{c}{\textbf{Centre-Embedding}} &
\multicolumn{2}{c}{\textbf{WSQ1}} &
\multicolumn{2}{c}{\textbf{WSQ2}} \\
\cmidrule(lr){3-4}\cmidrule(lr){5-6}\cmidrule(lr){7-8}\cmidrule(lr){9-10}\cmidrule(lr){11-12}
 &  & \textbf{rep\_nr} & \textbf{ran\_nr} & \textbf{rep\_nr} & \textbf{ran\_nr} &
      \textbf{rep\_nr} & \textbf{ran\_nr} & \textbf{rep\_nr} & \textbf{ran\_nr} &
      \textbf{rep\_nr} & \textbf{ran\_nr} \\
\midrule
\multirow{3}{*}{[0, 0.2]} 
  & normal & 0.0689 & -0.0585 & 0.0723 & -0.0583 & 0.0530 & -0.0556 & 0.0427 & -0.0607 & 0.0408 & -0.0610 \\
  &induction & 0.0606 & -0.0568 & 0.0595 & -0.0566 & 0.0530 & -0.0556 & 0.0427 & -0.0607 & 0.0408 & -0.0610 \\
  & random & 0.0688 & -0.0569 & 0.0697 & -0.0569 & 0.0575 & -0.0550 & 0.0429 & -0.0587 & 0.0407 & -0.0590 \\
\addlinespace[2pt]
\multirow{3}{*}{[0.4, 0.6]} 
  & normal & 0.2604 & -0.1175 & 0.2636 & -0.1166 & 0.2200 & -0.1136 & 0.2148 & -0.1163 & 0.2128 & -0.1163 \\
  &induction & 0.2330 & -0.1149 & 0.2413 & -0.1142 & 0.2250 & -0.1120 & 0.2176 & -0.1150 & 0.2158 & -0.1149 \\
  & random & 0.2465 & -0.1151 & 0.2515 & -0.1137 & 0.2330 & -0.1104 & 0.2192 & -0.1134 & 0.2192 & -0.1131 \\
\addlinespace[2pt]
\multirow{3}{*}{[0.8, 1.0]} 
  & normal & 1.4750 & -0.1311 & 1.4879 & -0.1322 & 1.2582 & -0.1153 & 1.2463 & -0.1087 & 1.2329 & -0.1099 \\
  &induction & 1.2685 & -0.11545 & 1.2825 & -0.1167 & 1.2287 & -0.1086 & 1.2139 & -0.1026 & 1.1972 & -0.1037 \\
  & random & 1.4239 & -0.1267 & 1.4425 & -0.1292 & 1.2390 & -0.1102 & 1.2304 & -0.1030 & 1.2217 & -0.1051 \\
\bottomrule
\end{tabular}
\caption{\textbf{Mean activation at the last prompt token by segment (LLama-3.1-8B, $K{=}250$).}
\enquote{normal}: no ablation; \enquote{induction}: ablate top 3\% induction heads; \enquote{random}: ablate 3\% random heads.
\enquote{rep\_nr}: top-$K$ repetition neurons in the segment; \enquote{ran\_nr}: $K$ random neurons from the same segment.}
\label{tab:apx_abhd_avgnr_l8}
\end{table*}

\begin{table*}[t]
\centering
\footnotesize
\setlength{\tabcolsep}{4.8pt}
\begin{tabular}{llrrrrrrrrrr}
\toprule
\multirow{2}{*}{\textbf{Segment}} & \multirow{2}{*}{\textbf{Cond.}} &
\multicolumn{2}{c}{\textbf{Repetition}} &
\multicolumn{2}{c}{\textbf{Recursion}} &
\multicolumn{2}{c}{\textbf{Centre-Embedding}} &
\multicolumn{2}{c}{\textbf{WQS1}} &
\multicolumn{2}{c}{\textbf{WSQ2}} \\
\cmidrule(lr){3-4}\cmidrule(lr){5-6}\cmidrule(lr){7-8}\cmidrule(lr){9-10}\cmidrule(lr){11-12}
 &  & \textbf{rep\_nr} & \textbf{ran\_nr} & \textbf{rep\_nr} & \textbf{ran\_nr} &
      \textbf{rep\_nr} & \textbf{ran\_nr} & \textbf{rep\_nr} & \textbf{ran\_nr} &
      \textbf{rep\_nr} & \textbf{ran\_nr} \\
\midrule
\multirow{3}{*}{[0, 0.2]}
  & normal & -0.0060 & -0.0288 & -0.0288 & -0.0293 & -0.0104 & -0.0288 & -0.0252 & -0.0219 & -0.0248 & -0.0209 \\
  & induction   & -0.0065 & -0.0288 & -0.0288 & -0.0293 & -0.0109 & -0.0288 & -0.0257 & -0.0217 & -0.0254 & -0.0207 \\
  & random   & -0.0058 & -0.0283 & -0.0283 & -0.0286 & -0.0101 & -0.0281 & -0.0254 & -0.0220 & -0.0250 & -0.0212 \\
\addlinespace[2pt]
\multirow{3}{*}{[0.4, 0.6]}
  & normal &  0.2707 & -0.1049 &  0.1649 & -0.1066 &  0.2274 & -0.1021 &  0.2364 & -0.1195 &  0.2394 & -0.1186 \\
  & induction   &  0.2354 & -0.1023 &  0.1023 & -0.1047 &  0.1891 & -0.1002 &  0.1940 & -0.1187 &  0.1985 & -0.1179 \\
  & random   &  0.2544 & -0.1030 &  0.1030 & -0.1059 &  0.2132 & -0.1010 &  0.2217 & -0.1168 &  0.2247 & -0.1162 \\
\addlinespace[2pt]
\multirow{3}{*}{[0.8, 1.0]}
  & normal &  4.3277 & -0.0515 &  3.9553 & -0.0515 &  4.2318 & -0.0552 &  4.3482 & -0.0541 &  4.3472 & -0.0545 \\
  & induction   &  3.9102 & -0.0437 &  3.8471 & -0.0467 &  3.8461 & -0.0457 &  3.9265 & -0.0538 &  3.9449 & -0.0405 \\
  & random   &  4.2337 & -0.0471 &  4.0471 & -0.0493 &  4.1315 & -0.0526 &  4.2417 & -0.0477 &  4.2411 & -0.0499 \\
\bottomrule
\end{tabular}
\caption{\textbf{Qwen-2.5-7B: mean activation at the last prompt token by segment, $K{=}250$.}
\enquote{normal}: no ablation; \enquote{induction}: ablate top 3\% induction heads; \enquote{random}: ablate 3\% random heads.
\enquote{rep\_nr}: top-$K$ repetition neurons in the segment; \enquote{ran\_nr}: $K$ random neurons from the same segment.}
\label{tab:apx_abhd_avgnr_q7}
\end{table*}

\begin{table*}[t]
\centering
\footnotesize
\setlength{\tabcolsep}{4.8pt}
\begin{tabular}{llrrrrrrrrrr}
\toprule
\multirow{2}{*}{\textbf{Segment}} & \multirow{2}{*}{\textbf{Cond.}} &
\multicolumn{2}{c}{\textbf{Repetition}} &
\multicolumn{2}{c}{\textbf{Recursion}} &
\multicolumn{2}{c}{\textbf{Centre-Embedding}} &
\multicolumn{2}{c}{\textbf{WSQ1}} &
\multicolumn{2}{c}{\textbf{WSQ2}} \\
\cmidrule(lr){3-4}\cmidrule(lr){5-6}\cmidrule(lr){7-8}\cmidrule(lr){9-10}\cmidrule(lr){11-12}
 &  & \textbf{rep\_nr} & \textbf{ran\_nr} & \textbf{rep\_nr} & \textbf{ran\_nr} &
      \textbf{rep\_nr} & \textbf{ran\_nr} & \textbf{rep\_nr} & \textbf{ran\_nr} &
      \textbf{rep\_nr} & \textbf{ran\_nr} \\
\midrule
\multirow{3}{*}{[0, 0.2]}
  & normal & 0.1531 & -0.0443 & 0.1609 & -0.0437 & 0.1464 & -0.0439 & 0.1275 & -0.0525 & 0.1288 & -0.0520 \\
  & induction   & 0.1295 & -0.0442 & 0.1350 & -0.0436 & 0.1231 & -0.0439 & 0.1084 & -0.0522 & 0.1089 & -0.0517 \\
  & random   & 0.1507 & -0.0438 & 0.1577 & -0.0432 & 0.1436 & -0.0434 & 0.1213 & -0.0516 & 0.1288 & -0.0513 \\
\addlinespace[2pt]
\multirow{3}{*}{[0.4, 0.6]}
  & normal & 0.3769 & -0.0867 & 0.3828 & -0.0868 & 0.3617 & -0.0862 & 0.3998 & -0.0859 & 0.3825 & -0.0863 \\
  & induction   & 0.3733 & -0.0815 & 0.3780 & -0.0816 & 0.3676 & -0.0827 & 0.4026 & -0.0865 & 0.3891 & -0.0859 \\
  & random   & 0.4397 & -0.0842 & 0.4533 & -0.0834 & 0.4220 & -0.0840 & 0.4464 & -0.0842 & 0.4275 & -0.0839 \\
\addlinespace[2pt]
\multirow{3}{*}{[0.8, 1.0]}
  & normal & 0.5938 & -0.0764 & 0.5916 & -0.0769 & 0.5691 & -0.0794 & 0.6775 & -0.0736 & 0.6655 & -0.0732 \\
  & induction   & 0.4940 & -0.0649 & 0.4923 & -0.0654 & 0.4823 & -0.0668 & 0.5476 & -0.0683 & 0.5385 & -0.0662 \\
  & random   & 0.5784 & -0.0717 & 0.5860 & -0.0723 & 0.5559 & -0.0746 & 0.6319 & -0.0689 & 0.6204 & -0.0683 \\
\bottomrule
\end{tabular}
\caption{\textbf{Llama-2-13B: mean activation at the last prompt token by segment/condition, $K{=}250$.}
\enquote{normal}: no ablation; \enquote{induction}: ablate top 3\% induction heads; \enquote{random}: ablate 3\% random heads.
\enquote{rep\_nr}: top-$K$ repetition neurons in the segment; \enquote{ran\_nr}: $K$ random neurons from the same segment.}
\label{tab:apx_abhd_avgnr_l13}
\end{table*}

\section{Joint Ablation}
\label{apx:joint_ab}

\begin{figure}
    \centering
    \includegraphics[width=\columnwidth]{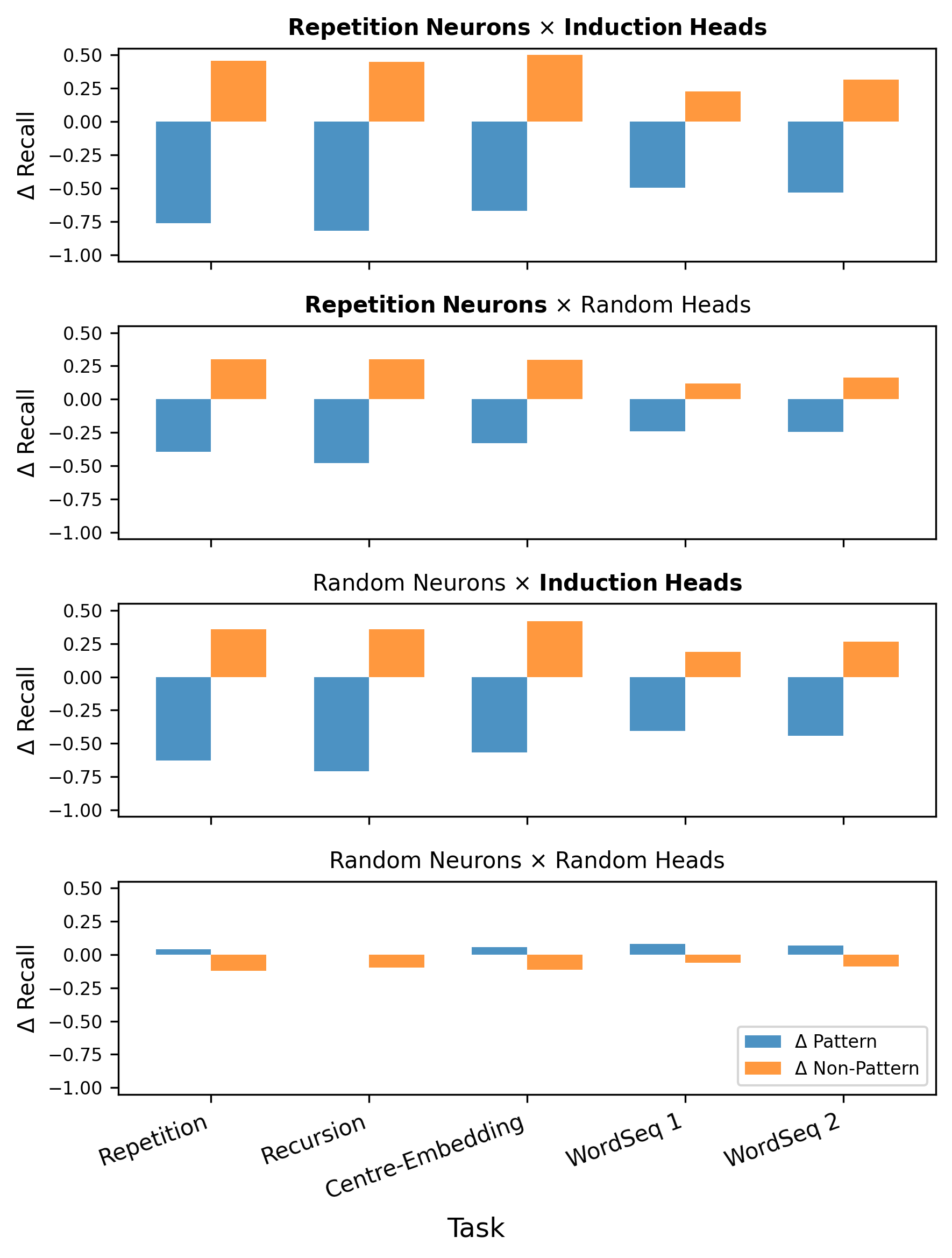}
    \caption{Change in LlaMA-3.1-8B recall (\(\Delta\)Recall) for Pattern (blue) and Non-Pattern (orange) inputs on five tasks—rep, rec, ceb, wsq1, wsq2—under four joint‐ablation regimes: (a) repetition neurons(top 250 from final segment [0.8–1.0]) \(\times\) induction heads (top 3\% by prefix-matching score), (b) repetition neurons \(\times\) random heads, (c) random neurons \(\times\) induction heads, and (d) random neurons \(\times\) random heads with 5-shot}
    \label{fig:llama318-jointab}
\end{figure}

\begin{figure}
    \centering
    \includegraphics[width=\columnwidth]{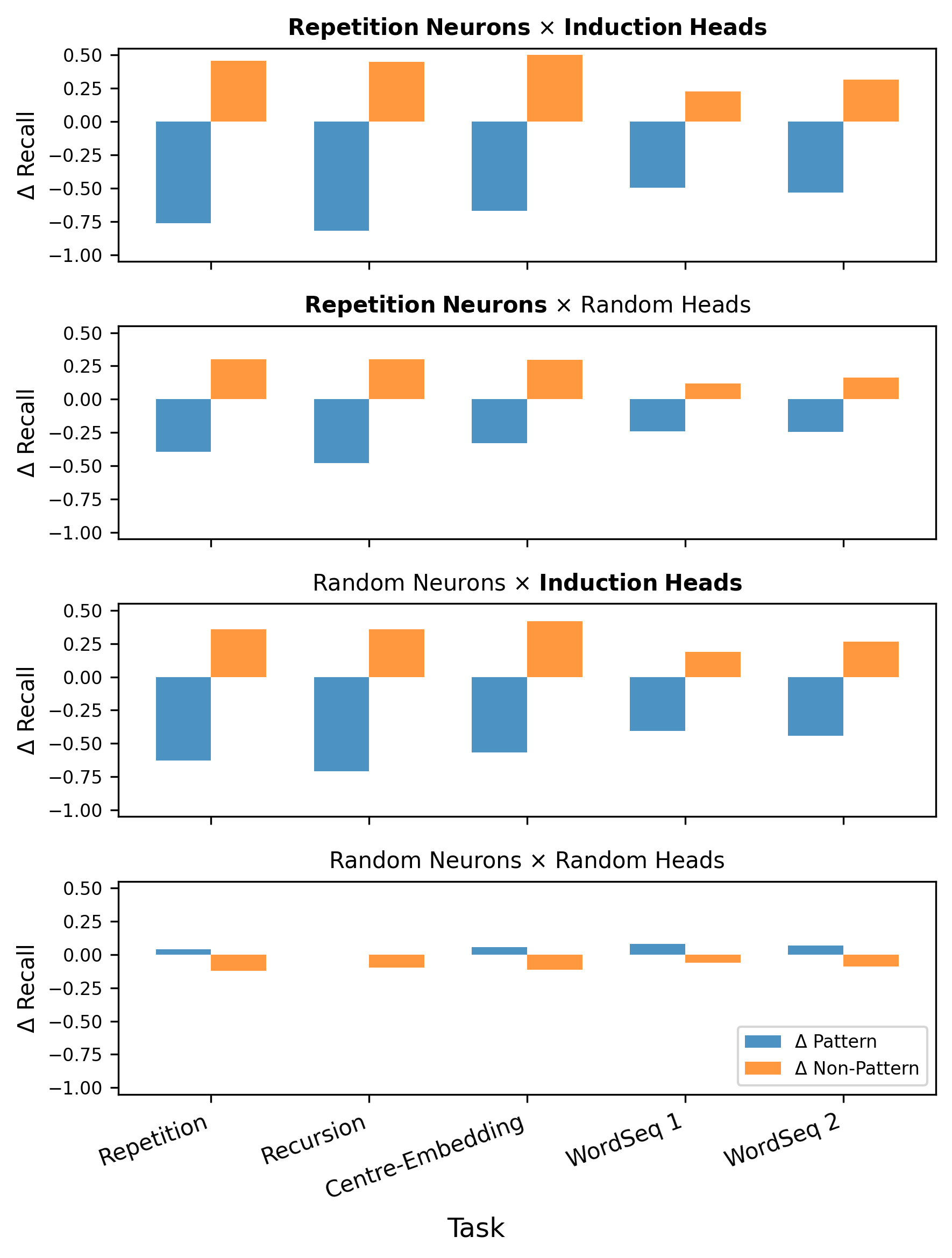}
    \caption{Change in Qwen2.5-7B recall (\(\Delta\)Recall) for Pattern (blue) and Non-Pattern (orange) inputs on five tasks—rep, rec, ceb, wsq1, wsq2—under four joint‐ablation regimes: (a) repetition neurons(top 250 from final segment [0.8–1.0]) \(\times\) induction heads (top 3\% by prefix-matching score), (b) repetition neurons \(\times\) random heads, (c) random neurons \(\times\) induction heads, and (d) random neurons \(\times\) random heads with 10-shot}
    \label{fig:enter-label}
\end{figure}

\begin{figure}
    \centering
    \includegraphics[width=\columnwidth]{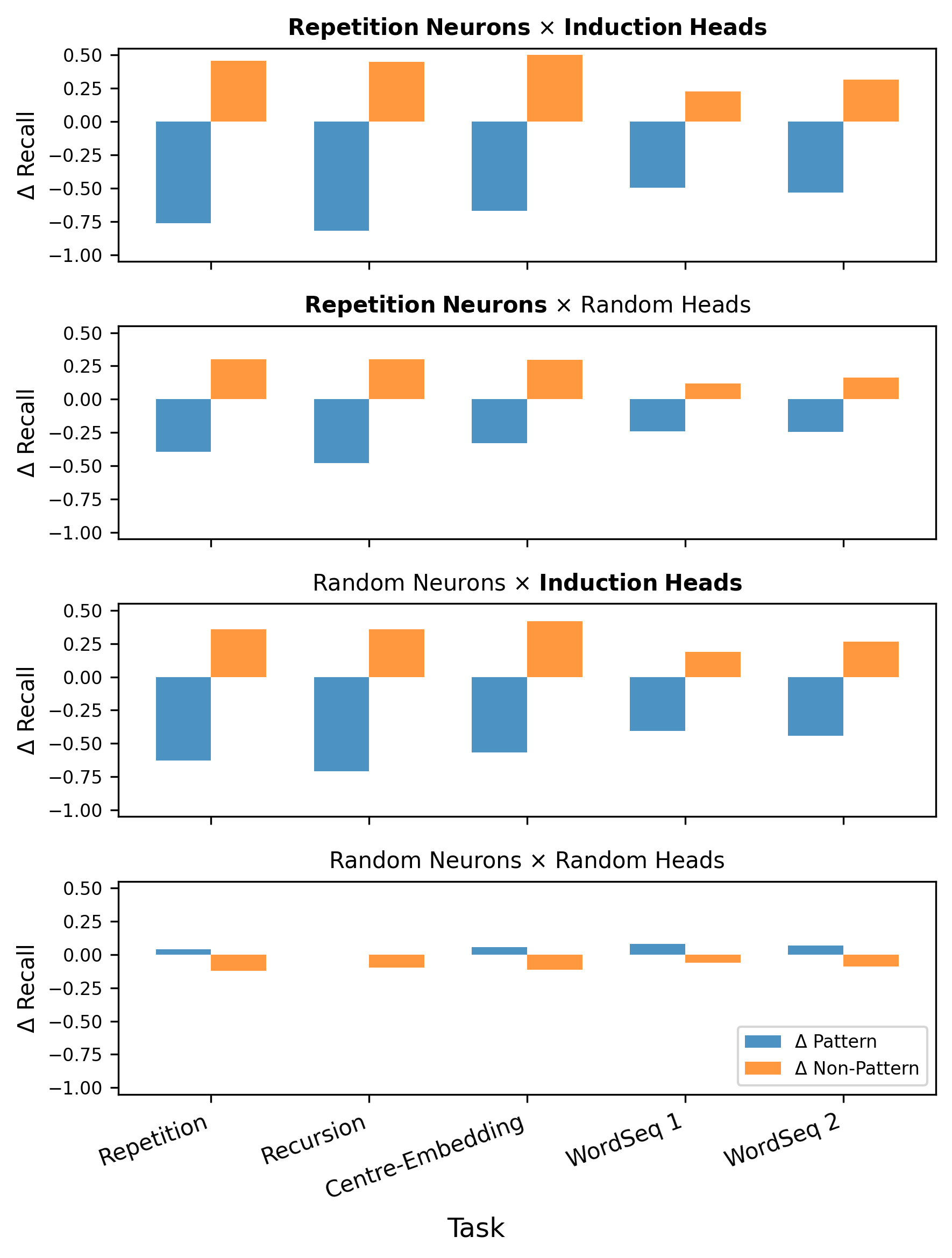}
    \caption{Change in LlaMA-2-13B recall (\(\Delta\)Recall) for Pattern (blue) and Non-Pattern (orange) inputs on five tasks—rep, rec, ceb, wsq1, wsq2—under four joint‐ablation regimes: (a) repetition neurons(top 250 from final segment [0.8–1.0]) \(\times\) induction heads (top 3\% by prefix-matching score), (b) repetition neurons \(\times\) random heads, (c) random neurons \(\times\) induction heads, and (d) random neurons \(\times\) random heads with 10-shot}
    \label{fig:llama13-jointab}
\end{figure}

Results for the other models are illustrated in Figures~\ref{fig:llama318-jointab}–\ref{fig:llama13-jointab}. Across all settings, ablating both repetition neurons and induction heads consistently reduces the model’s performance most severely on the Pattern class. This confirms the sequential cascade described in the main sections.

\section{Text Generation Analysis}
\label{apx:text_gen}
\begin{figure}
    \centering
    \includegraphics[width=1\linewidth]{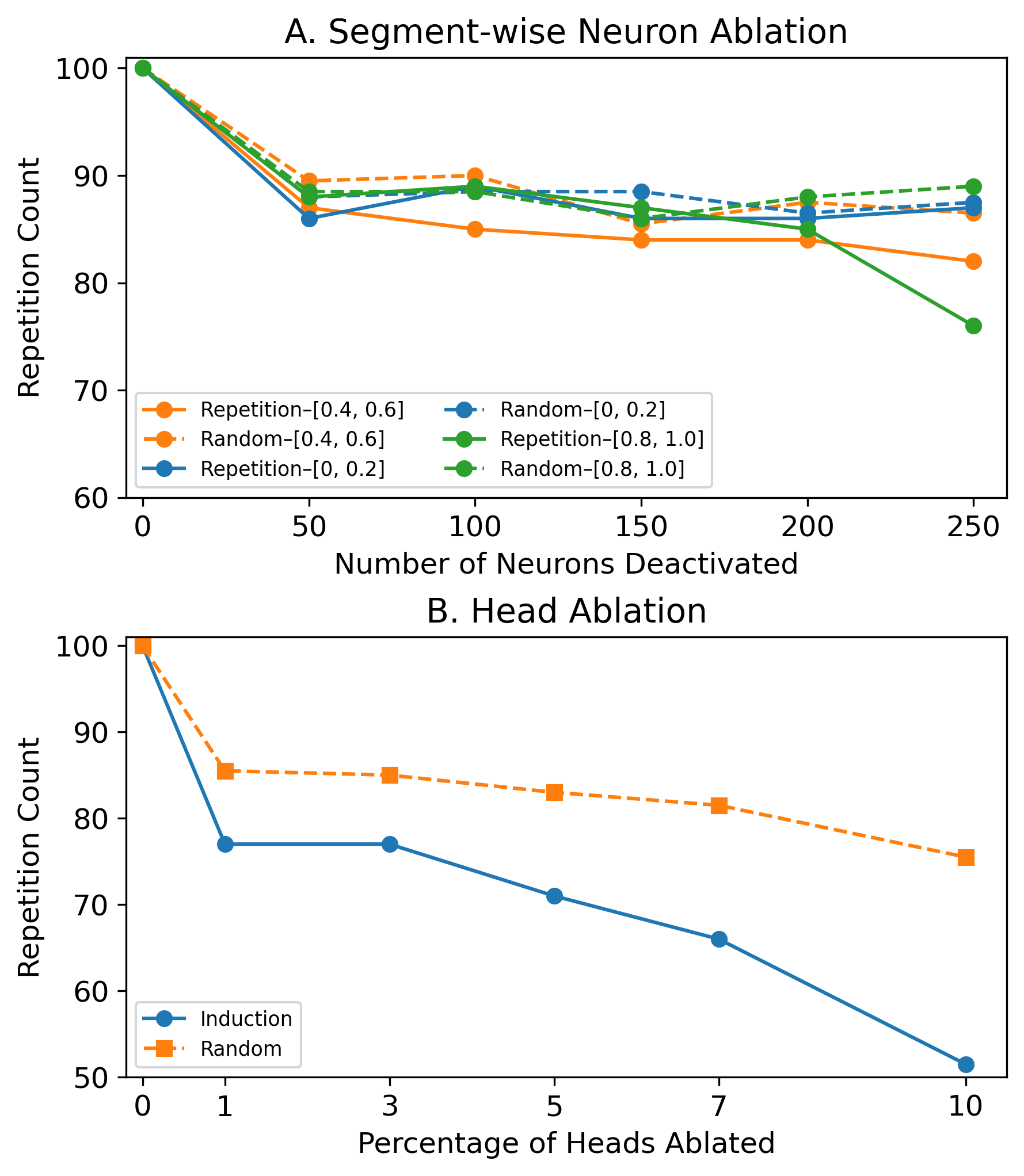}
    \caption{Effect of ablating repetition neurons and attention heads on generated repetition text in Qwen-2.5-7B}
    \label{fig:x}
\end{figure}

\begin{figure}
    \centering
    \includegraphics[width=1\linewidth]{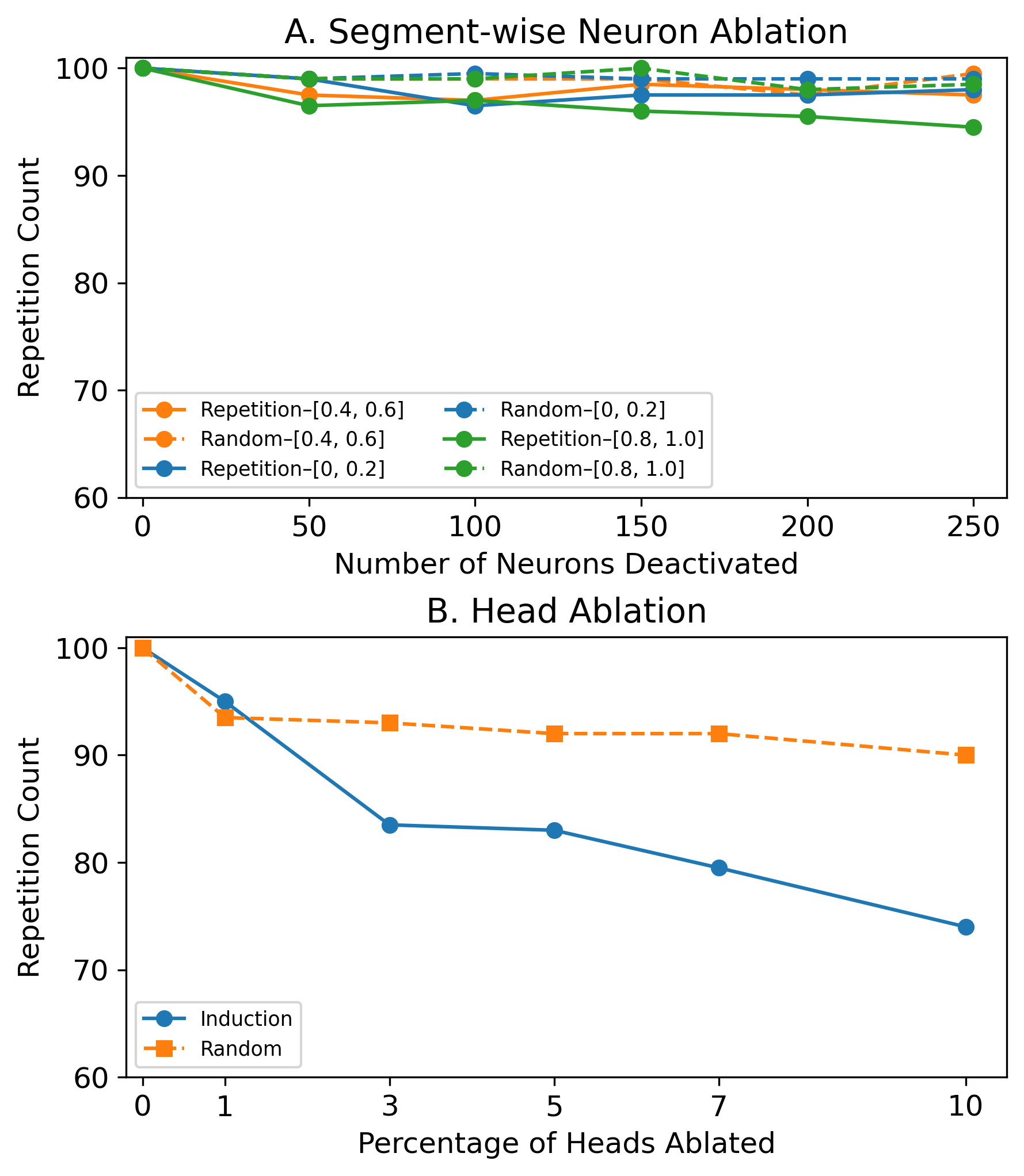}
    \caption{Effect of ablating repetition neurons and attention heads on generated repetition text in LlaMA-2-13B}
    \label{fig:y}
\end{figure}

Figures \ref{fig:x} and \ref{fig:y} show the results of ablations on induction heads and repetition neurons in our repetitive-generation experiments.


\textbf{Downstream Tasks} As in Table \ref{tab:sst2-wmt-sample}, we use Positive and Negative as label for the SST2. In 0-shot setting, we add the instruction for label as \textit{Positive or Negative} for SST2 and \textit{to English} for the WMT19. We also provide detail results on SST2 and WMT19 from Tables \ref{tab:sst2-rep-ablation} to \ref{tab:wmt19-last}.

\begin{table*}[ht]
  \centering
  \small
  \caption{Few-shot prompts and queries for SST2 and WMT19 tasks}
  \label{tab:sst2-wmt-sample}
  \begin{tabular}{@{} l l p{3.2cm} p{5.5cm} l p{1cm}}
    \toprule
    \textbf{Task} & \textbf{Shots} & \textbf{Query} & \textbf{Prompt} & \textbf{GT} \\
    \midrule

    \multirow{2}{*}{SST2}
      & 0  & \texttt{"that loves its characters and communicates something rather beautiful about human nature"}  
        & Classify the review according to its sentiment (Positive or Negative). \textbf{Review: } $<query>$ \textbf{Sentiment: } 
        & Positive
    \\

      & 5 
        & \texttt{"uneven film"}  
        & Classify the review according to its sentiment (Positive or Negative). \textbf{Review: } $<demo\_1>$ \textbf{Sentiment: $<label\_1>$} ...$(5-shot)$ \textbf{Review: } $<query>$ \textbf{Sentiment: }
        & Negative
    \\[1ex]
    \hdashline
    \multirow{2}{*}{WMT19}
      & 0  & \texttt{"Wiederaufnahme der Sitzungsperiode"}  
        & Below is a list of various sequences. Your task is to translate from source language to English as target language . \textbf{Source: } $<query>$ \textbf{Target: }
        & Resumption of the session
    \\
    
      & 5 & \texttt{"Vielen Dank, Herr Barón Crespo."}  
        & Below is a list of various sequences. Your task is to translate from source language to English as target language . \textbf{Source: } $<demo\_1>$ \textbf{Target: $<label\_1>$} ...$(5-shot)$ \textbf{Source: } $<query>$ \textbf{Target: }
        & Thank you, Mr Barón Crespo.
    \\[1ex]

    \bottomrule
  \end{tabular}
\end{table*}
\begin{table*}[t]
\centering
\small
\begin{tabular}{l c c c c c}
\toprule
Model & Shot & Full model & Segment [0.0,0.2] & Segment [0.4,0.6] & Segment [0.8,1.0] \\
\midrule
LLaMA-3.1-8B   & 0 & 0.891 & 0.888 & 0.890 & 0.894 \\
LLaMA-3.1-8B   & 5 & 0.889 & 0.891 & 0.896 & 0.857 \\
Qwen-2.5-7B    & 0 & 0.948 & 0.942 & 0.952 & 0.933 \\
Qwen-2.5-7B    & 5 & 0.935 & 0.933 & 0.937 & 0.932 \\
LLaMA-2-13B    & 0 & 0.810 & 0.819 & 0.831 & 0.839 \\
LLaMA-2-13B    & 5 & 0.942 & 0.940 & 0.938 & 0.937 \\
\bottomrule
\end{tabular}
\caption{F1-score of ablation of 250 repetition neurons in SST-2 (1,000 samples, 500 per class). Ablating repetition neurons from middle layers usually maintains performance, while those from last layers are less stable.}
\label{tab:sst2-rep-ablation}
\end{table*}
\begin{table*}[t]
\centering
  \scriptsize
  \setlength{\tabcolsep}{2pt} 
  \resizebox{\linewidth}{!}{
  \begin{tabular}{@{}l c
    ccc   ccc   ccc   ccc@{}}
    \toprule
    & & \multicolumn{3}{c}{BLEU} 
        & \multicolumn{3}{c}{ROUGE-L} 
        & \multicolumn{3}{c}{Distinct-2} 
        & \multicolumn{3}{c}{BERTScore (F1)} \\
    Model        & Shot 
      & (1) & (2) & (3) 
      & (1) & (2) & (3) 
      & (1) & (2) & (3) 
      & (1) & (2) & (3) \\
    \midrule
    LLaMA-3.1-8B & 0  & -0.804 & 0.185  & 0.047  & -0.186 & 0.080  & 0.300  & -0.004 & -0.003 & -0.005 & -0.003 & -0.001 & 0.000 \\
    LLaMA-3.1-8B & 5  & -0.622 & -0.164 & -0.199 & -0.552 & 0.008  & 0.182  & -0.005 & -0.003 & -0.003 & -0.001 & 0.000  & 0.000 \\
    Qwen-2.5-7B  & 0  & 0.976  & 0.356  & 0.620  & 5.601  & 0.260  & 0.744  & 0.040  & 0.037  & 0.021  & 0.004  & -0.002 & 0.001 \\
    Qwen-2.5-7B  & 5  & -0.465 & 0.412  & 0.405  & -0.369 & 0.192  & 0.220  & -0.001 & -0.008 & 0.001  & -0.001 & 0.000  & -0.001 \\
    LLaMA-2-13B  & 0  & 0.969  & -0.495 & 1.726  & 1.034  & -0.274 & 2.924  & 0.001  & 0.003  & -0.006 & 0.005  & 0.000  & 0.012 \\
    LLaMA-2-13B  & 5  & 0.031  & -0.186 & -0.451 & -0.442 & -0.194 & -0.585 & 0.000  & 0.002  & 0.006  & 0.000  & 0.000  & 0.000 \\
    \bottomrule
  \end{tabular}
  }
  \caption{Change in model performance (after–before) when ablating 250 repetition neurons in middle-layers [0.4,0.6] on WMT19 (1050 samples: cs-en (1), de-en (2), zh–en (3)).}
  \label{tab:wmt19-mid}
\end{table*}

\begin{table*}[t]
  \centering
  \resizebox{\linewidth}{!}{%
  \begin{tabular}{@{}l c ccc ccc ccc ccc@{}}
    \toprule
    & & \multicolumn{3}{c}{BLEU} & \multicolumn{3}{c}{ROUGE-L} 
        & \multicolumn{3}{c}{Distinct-2} & \multicolumn{3}{c}{BERTScore (F1)} \\
    Model        & Shot 
      & (1) & (2) & (3)
      & (1) & (2) & (3)
      & (1) & (2) & (3)
      & (1) & (2) & (3) \\
    \midrule
    LLaMA-3.1-8B & 0 & -12.048 & -6.747 & -5.810  
                 & -12.859 & -8.905 & -4.561  
                 & -0.010  & -0.060 & -0.070  
                 & -0.049  & -0.036 & -0.017 \\
    LLaMA-3.1-8B & 5 &  -4.033 & -3.552 & -8.249  
                 &  -2.409 & -3.915 & -7.532  
                 & -0.013  & -0.042 & -0.065  
                 & -0.004  & -0.006 & -0.013 \\
    Qwen-2.5-7B  & 0 &   0.660 &  2.899 &  2.055  
                 &   3.030 &  5.196 &  3.076  
                 & -0.001  &  0.105 &  0.083  
                 &  0.005  &  0.013 &  0.007 \\
    Qwen-2.5-7B  & 5 &  -0.500 & -0.591 &  0.368  
                 &  -0.044 & -0.569 & -0.139  
                 &  0.029  &  0.018 &  0.028  
                 & -0.002  & -0.001 &  0.002 \\
    LLaMA-2-13B  & 0 &   0.772 & -0.027 &  0.886  
                 &   3.593 & -0.174 &  0.762  
                 &  0.000  & -0.007 & -0.060  
                 &  0.003  & -0.003 &  0.013 \\
    LLaMA-2-13B  & 5 &   0.269 &  0.257 & -0.821  
                 &  -0.356 & -0.072 & -0.891  
                 & -0.007  & -0.011 & -0.007  
                 & -0.001  &  0.000 & -0.001 \\
    \bottomrule
  \end{tabular}
  }
  \caption{Change in model performance (after–before) when ablating 250 repetition neurons in last‐segment layers [0.8,1.0] on WMT19 (1,050 samples: cs–en (1), de–en (2), zh–en (3)).}
\label{tab:wmt19-last}
\end{table*}

\end{document}